\documentclass[prx,nofootinbib,onecolumn,showkeys,groupaddress,preprintnumbers,floatfix,
superscriptaddress]{revtex4-1}

\usepackage{dynlearn}
\usepackage{multirow}
\fxsetup{status=draft}
\usepackage{empheq}
\usepackage{overpic}
\usepackage{listings}
\usepackage{hyperref}
\usepackage{url}
\usepackage{xstring}

\newcommand{\argmin}{\text{argmin}}

\newcommand{\SSet}{\boldsymbol{\mathcal{S}}}
\newcommand{\Abet}{\mathcal{X}}

\newcommand{\stationary}{\boldsymbol{\pi}}

\newcommand{\dbra}[1]{\bra{ \! \langle #1}}
\newcommand{\dket}[1]{\ket{ #1 \rangle \!}}
\newcommand{\dbraket}[1]{\braket{ \! \langle #1 \rangle \!}}
\newcommand{\init}{\dbra{\eta^{(\varnothing)}}}

\newcommand{\act}[1][w]{\vec{a}_{#1}} 
\newcommand{\genBloch}[1][w]{\vec{b}_{#1}}

\newcommand{\params}{\boldsymbol{\theta}}

\newcommand{\tr}{\text{tr}}

\newcommand{\sysref}{I/d}  
\newcommand{\constname}{\xi}

\makeatletter
\@booleanfalse\titlepage@sw
\makeatother

\begin{document}

\def\ourTitle{%
Neural networks leverage 
nominally quantum and post-quantum representations
}

\def\ourAbstract{%
We show that deep neural networks, including transformers and RNNs, pretrained as usual on next-token prediction, intrinsically discover and 
represent beliefs over
`quantum' and `post-quantum' low-dimensional generative models of their training data---as if performing iterative Bayesian updates over the latent state of this world model during inference as they observe more context.
Notably, neural nets easily find these representation whereas there is no finite classical circuit that would do the job. 
The corresponding geometric relationships among neural activations induced by different input sequences are found to be largely independent of neural-network architecture. 
Each point in this geometry corresponds to a history-induced probability density over all possible futures, and the relative displacement of these points reflects
the 
difference in mechanism and magnitude 
for how these distinct pasts affect the future.
%
%
%
%
%
}

\def\ourKeywords{%
}

\hypersetup{
  pdfauthor={Paul M. Riechers},
  pdftitle={\ourTitle},
  pdfsubject={\ourAbstract},
  pdfkeywords={\ourKeywords},
  pdfproducer={},
  pdfcreator={}
}


\title{\ourTitle}

\author{Paul M.\ Riechers}
\email{pmriechers@gmail.com}

\affiliation{Simplex,
	Astera Institute,
	Emeryville, CA}

\affiliation{Beyond Institute for Theoretical Science (BITS),
San Francisco, CA}

\author{Thomas J.\ Elliott} 

\affiliation{
	Department of Physics \& Astronomy, University of Manchester, Manchester M13 9PL, United Kingdom }
	
\affiliation{
	Department of Mathematics, University of Manchester, Manchester M13 9PL, United Kingdom
}
	
\author{Adam S.\ Shai}

\affiliation{Simplex, Astera Institute,
	Emeryville, CA}
	

%
%
%
%
%
%


\date{\today}

\begin{abstract}
\ourAbstract
\end{abstract}




\date{\today}
\maketitle


\setstretch{1.1}



\section{Introduction}

On their surface,
the approaches of 
theoretical physics and machine learning 
couldn't be more different.
The first applies parsimonious principled theory to understand the nature of the world around us, while the latter throws a black box at the problem and captures structure in whatever messy way 
it can. 
Yet here we discover much more similarity than 
one would previously expect.  It turns out that deep neural networks too
find
parsimonious descriptions of data---evidently discovering nominally quantum and post-quantum representations of classical stochastic processes, which allow predictions of intricate correlations with vastly fewer dimensions than would otherwise be 
deemed necessary
in the naive characterization of a classical computational theory.

In the following, 
we demonstrate 
\emph{geometric representations of beliefs} 
linearly embedded
in the activations of neural networks 
that 
(i) 
spontaneously emerge over the course of standard 
self-supervised 
next-token-prediction
pretraining
(ii)
are 
universal across
deep neural-network architectures 
(including RNNs, LSTMs, and transformers),
(iii)
can be anticipated by theory,
and
(iv) 
directly
embody 
nominally quantum and post-quantum 
encodings for 
memory compression.
Given this universality, 
we anticipate that similar memory compression techniques are 
already implicitly leveraged by many trained models, including frontier AI models 
that increasingly affect the trajectory of humanity.
By demonstrating both universality and post-classical 
memory compression,
these advances
significantly 
expand on
our recent discovery that 
the neural activations of 
transformers 
pretrained on next-token prediction
linearly represent 
(sometimes fractal)
geometries of 
beliefs about the entire future~\cite{Shai24_Transformers, Piotrowski25_Constrained}.

This likely sounds paradoxical:
How can artificial neural networks---which are by all accounts classical, in the sense of ``understandable and implementable by classical (i.e., pre-quantum) physics''---utilize `quantum' and even `post-quantum' memory compression advantages 
to surpass the abilities of 
any classical 
computing 
system?
To precisely frame our results
and resolve this apparent paradox
requires 
that we recall the standard paradigms for classical memory and quantum memory, 
and how these memories are used in the standard frameworks for 
logical gate-based
computation.

\begin{figure}
    \centering
    \includegraphics[width=1\linewidth]{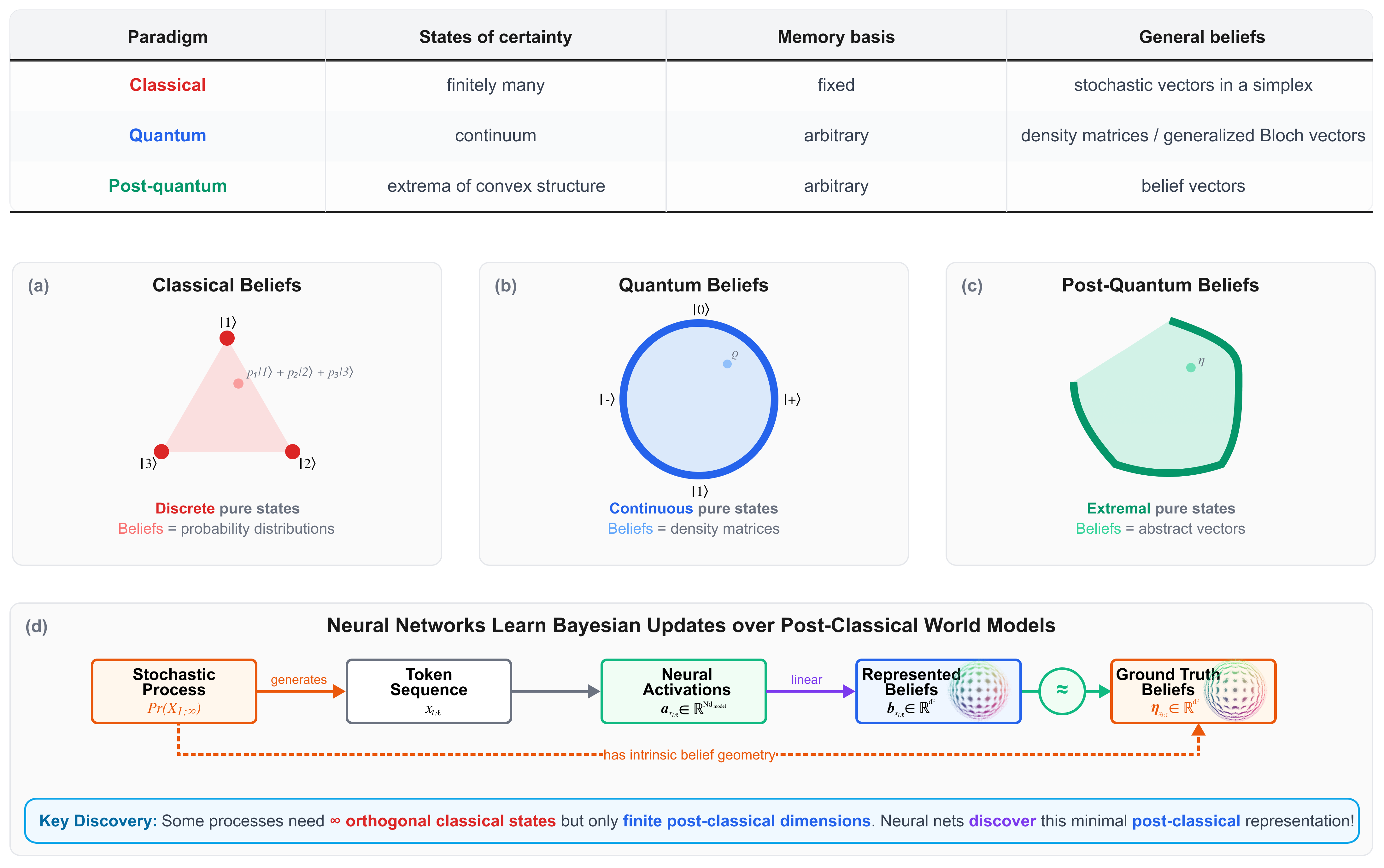}
\caption{\textbf{From classical to quantum to post-quantum belief representations: Neural networks learn to perform Bayesian inference over post-classical world models.} (a-c) Three computational paradigms of increasing generality for representing beliefs about stochastic processes. Classical beliefs (a) are probability distributions confined to simplices with discrete pure states (red vertices). Quantum beliefs (b) form density matrices with pure states on a continuous manifold (blue circle boundary). Post-quantum beliefs (c) have pure states that are extremal points of general convex sets (dark green)---these extremal points cannot be represented as convex combinations of other states. In each case, mixed states (interior regions) represent uncertainty. (d) How neural networks discover these representations. A stochastic process $\Pr(X_{1:\infty})$ generates token sequences $x_{1:\ell} \in \mathcal{X}^\ell$ and has intrinsic belief geometry. Neural networks trained on these sequences learn to generate activations that, through an affine map $\mathcal{L}$, encode belief states matching the ground truth geometry ($\approx$). Key discovery: Some processes requiring infinite classical states need only finite post-classical dimensions (e.g., 3D for a single qubit). 
After training on a classical stochastic process with a reduced quantum model with Hilbert-space dimension $d$,
we find that there is a fixed affine map $\mathcal{L}$ from any activation pattern 
$\act[x_{1:\ell}] \in \mathbb{R}^{N d_\text{model}}$
(across the $N$ layers of the neural network at token position $\ell$)
to the corresponding generalized Bloch vector $\genBloch[x_{1:\ell}] \in \mathbb{R}^{d^2-1}$ that would be obtained from Bayesian updates to the qudit memory; equivalently, there is another fixed affine map $\mathcal{L}'$ from activation patterns to the Kraus-updated density matrix $\rho_{x_{1:\ell}} \in \mathbb{C}^{d \times d}$.
Neural networks automatically discover these minimal quantum or post-quantum representations during standard next-token prediction training, despite having no explicit knowledge of the underlying generative model. This demonstrates that neural networks transcend limits of classical computational models by leveraging their continuous activation spaces to implicitly perform Bayesian inference over post-classical world models.}
\label{fig:overview}
\end{figure}




The results of this investigation yield
a deeper understanding of the true nature of computation in neural networks.
We find
that neural networks are not restricted to implementing classical computational circuits \`a la discrete logic gates.  Underneath the hood, they emulate an altogether different and sometimes more powerful mode of computation. 

Some classical stochastic processes can be generated via a single qubit of quantum memory, while the minimal classical memory would require infinitely many classical bits.
To predict such a process, one could perform Bayesian updates over a low-dimensional quantum world model---we reveal that standard neural networks learn this low-dimensional representation from pretraining as usual on next-token prediction.

\subsection{Classical, quantum, and post-quantum computing paradigms}

Distinct classical memory states 
of a digital computer 
are 
mutually exclusive by design (Fig.~\ref{fig:overview}a).
This orthogonal 
set of possible memories can be thought of as a partition of a more refined set of orthonormal microstates of a physical memory system
\cite{Riec19_Transforming}.\footnote{E.g., there are many physical microstates consistent with the logical $\texttt{001}$ memory state of three bits of magnetic memory. The specific microstate within each equivalence class is by design irrelevant to the logical operations.}
With the further allowance of subjective uncertainty over these memory states,
the most general computationally relevant `state' of a classical memory is a probability distribution, living in the probability simplex over these memory states, which is the convex hull of these orthogonal pure states of certainty.  
The state of a classical memory system
can thus be identified as a probability vector, with non-negative vector elements summing to unity.
Classical memory lives in a probability simplex.

Quantum memory allows more flexibility:
quantum superposition avails a continuum of pure states, 
from linear combinations of a finite set of classical basis states (Fig.~\ref{fig:overview}b).
For a single qubit, this means that every point on the surface of the Bloch sphere is a valid pure state
$\ket{\psi} = c_0 \ket{0} + c_1 \ket{1}$ with $c_0, \, c_1 \in \mathbb{C}$ and $|c_0|^2 + |c_1|^2 = 1$,
whereas a classical bit
would only admit the two pure orthogonal computational basis states $\ket{0}$ and $\ket{1}$
that are represented as disconnected antipodal points on the north and south pole of the Bloch sphere as depicted in Fig.~\ref{fig:overview}b.
Quantum states can always be represented 
as positive semidefinite Hermitian operators with unit trace---the familiar density matrix.
This allows for all convex combinations of the pure states $\rho = \sum_n p_n \ket{\psi_n} \! \bra{\psi_n}$ with $p_n \in [0,1]$ and $\sum_n p_n = 1$.  For a single qubit, the nonpure mixed states fill in the Bloch ball~\cite{Nielsen10_Quantum}.

%

Hypothetical 
post-quantum 
generalized probabilistic
theories (GPTs) allow yet more freedom, 
in both the definition of states and their transformations,
notably allowing for stronger-than-quantum correlations in Bell-like tests of spatial nonlocality~\cite{Plavala23_General}.
The class of generalized probabilistic theories includes both quantum and classical theory as special cases, but also accommodates more general theories that may be natural candidates if quantum theory is eventually falsified. 
By relaxing some of the assumptions associated with quantum and classical theories, 
one can think of the state of a GPT quite generally as a vector in a convex bounded subset of a real finite-dimensional vector space (Fig.~\ref{fig:overview}c).

In all cases, \emph{pure states} are those states that cannot be obtained by convex combinations of other states---they correspond to states of certainty.


In the following, we are interested in 
\emph{stochastic processes}, where a stochastic process can be thought of as a probability density over all possible sequences of tokens.
Moreover, we are interested in the resources required to generate and predict any particular stochastic process, and whether the process has a 
finite-dimensional classical, quantum, or post-quantum description.
For example, a finite system using classical memory to generate a stochastic process must belong to the class of hidden Markov models (HMMs), since more general classical computing concepts like stacks and tapes in principle require access to an unbounded number of degrees of freedom.
A finite system using quantum memory may have access to some number of qubits, 
and quantum instruments that implement 
general updates to the quantum memory while reporting the classical measurement readouts that correspond to observable tokens.






Contemporary literature defines three classes of stochastic processes:
\begin{enumerate}
	\item
	$\mathcal{C}$: The set of processes that can be simulated by a 
    classical
    finite-state hidden Markov model (HMM);\footnote{The restriction to a finite number of states is necessary for the whole theory to remain nontrivial, since all processes can in principle be simulated by an infinite-state HMM.}
	\item
	$\mathcal{Q}$: The set of processes that can be simulated by a quantum system with a finite-dimensional Hilbert space;
	\item
	$\mathcal{G}$: The set of processes that can be simulated by a finite-dimensional generalized probabilistic theory (GPT).
\end{enumerate}

It has been shown that $\mathcal{C} \subsetneq \mathcal{Q} \subsetneq \mathcal{G}$~\cite{Fanizza24_Quantum}.
In other words, finite quantum systems can simulate classical stochastic processes that would require infinitely many `classical' memory states in the smallest HMM~\cite{Monras16_Quantum}. 
Moreover, Fanizza et al.~\cite{Fanizza24_Quantum} recently established that finite-dimensional generalized probabilistic theories 
can generate classical stochastic processes that 
no finite-dimensional quantum system can.

\begin{figure}
    \centering
    \includegraphics[width=0.975\linewidth]{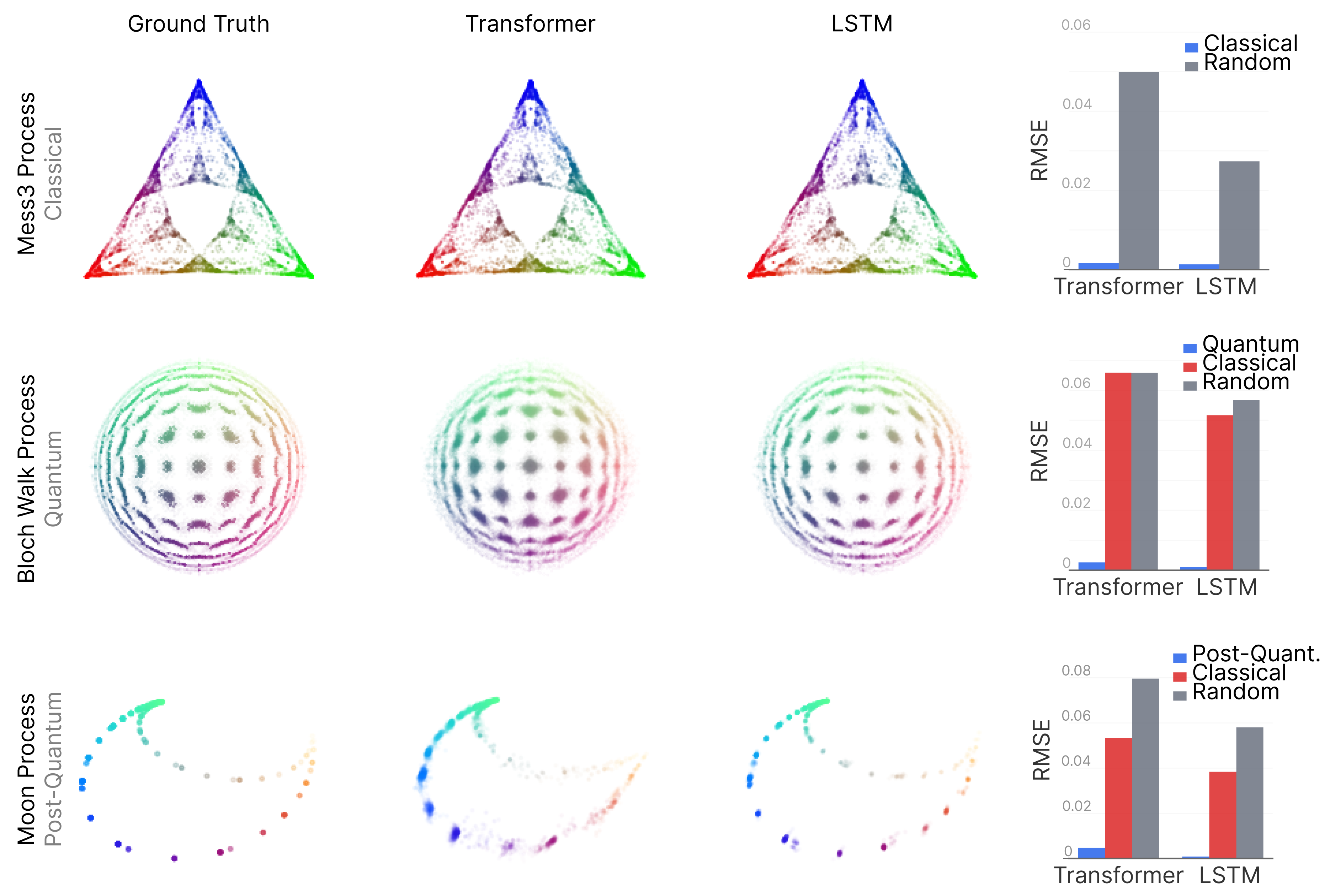}
    \caption{\textbf{Neural networks discover and linearly represent the minimal belief geometry of their training data.} Three rows show different stochastic processes requiring increasingly exotic minimal representations. \textbf{Top row (Mess3 Process):} A classical process with a 3-state HMM generator, where beliefs form a fractal pattern within the 2-simplex of probability distributions over the latent states of the generator. \textbf{Middle row (Bloch Walk Process):} A quantum process requiring a single qubit, with beliefs forming a 2D slice of the Bloch sphere---no finite classical HMM can generate this process. \textbf{Bottom row (Moon Process):} A post-quantum process that cannot be generated by any finite-dimensional quantum system. For each process: (left) ground truth belief geometry from exact Bayesian filtering over the minimal generator; (center) affine mapping of the activations in transformers to the ground truth beliefs predictions; (right) affine mapping of LSTM activations. Points are colored by position in ground-truth space and weighted by occurrence probability. Bar plots show root mean square error (RMSE) for fits to: the true minimal generator (blue), the best Markov-order-3 finite classical approximation (red), and using randomly initialized networks (gray). Networks learn to structure their activations according to the minimal representation---whether classical, quantum, or post-quantum---despite having no explicit knowledge of the generator structure. The high error for random networks confirms these geometries are learned through training, and are not architectural artifacts.}
    \label{fig:main_results}
\end{figure}

\subsection{Neural networks transcend the classical, quantum, and post-quantum distinction}

Notably, these previous works showing $\mathcal{C} \subsetneq \mathcal{Q} \subsetneq \mathcal{G}$ did not rely on the tensor-product compositional nature of quantum or post-quantum subsystems but, rather, leveraged the 
possible nonorthogonality of distinct pure states and the
greater flexibility the states have to transform in a finite-dimensional real vector space.\footnote{Although density matrices have matrix representations with complex-valued elements, the vector space of self-adjoint linear operators (i.e., Hermitian operators) is a real vector space.  (Recall that their eigenvalues are real, and so linear combinations of these are restricted to real coefficients to remain in the space of self-adjoint operators.)}
However, a moment of thought reveals that modern artificial neural networks provide an effectively real-valued vector space for activations (up to floating-point machine precision).\footnote{Perhaps our arguments apply to biological neural networks too, but we don't investigate this tantalizing possibility here.}
Hence, it at least seems plausible that neural networks may be able to encapsulate the memory-dimension advantage of quantum and post-quantum frameworks.
Indeed, in the following we demonstrate that 
\emph{deep neural networks pretrained on next-token prediction learn 
not only quantum and post-quantum generative models of their training data but, moreover, represent and update Bayesian beliefs over these exotic states as tokens are sequentially observed in context} (Fig.~\ref{fig:overview}d and Fig.~\ref{fig:main_results}).


Strikingly, 
the middle row of Fig.~\ref{fig:main_results} shows
an example where the neural activations of a standard transformer neural network directly (i.e., linearly) 
map to 
points in the Bloch sphere, 
jumping from one point in the Bloch sphere to another as more tokens are observed,
exactly as one would expect if the neural network were performing Bayesian updates over the qubit
density matrix 
associated with a
compact quantum generative model of its training data.



Figure \ref{fig:main_results} shows that neural activations of pretrained neural networks directly instantiate the belief geometry associated with Bayesian updates to the latent states of classical, quantum, and post-quantum world models, moving up to the more generalized category whenever it allows a representation in lower dimensions.
Moreover, these representations are universal across different types of neural network architecture---whether RNNs, LSTMs, transformers, or something else---and are invariant to the choice of hyperparameters for the neural net architecture.
Whenever neural networks learn to predict future tokens well, they learn to represent belief geometry.

The remainder of this paper fills out the notation, theory, experiments, and discussion needed to appreciate the results in more detail.
Sec.~\ref{sec:GHMMs} introduces generalized hidden Markov models (GHMMs), which are general linear models 
that, with enough dimensions, are
capable of generating any discrete-time stochastic process.
In Sec.~\ref{sec:ConditionalProbs}, 
we discuss how to calculate joint probabilities of future events from these GHMMs, conditioned on past observations, and the implied geometry of beliefs about the future.  
Starting in Sec.~\ref{sec:NeuralNetsDiscoverGeometry}, we then show that neural network activations linearly represent these belief geometries, as if they perform Bayesian updates over the latent states of classical, quantum, or sometimes post-quantum models of their training data. 
These representations are universal across neural network architectures.

\section{Generalized hidden Markov models}
\label{sec:GHMMs}

The state space, dynamics, and observations from 
sequential measurements in 
generalized probabilistic theories,
including both classical and quantum processes as special cases,
can all be represented by
\emph{generalized hidden Markov models} (GHMMs).
These GHMMs are similar to those introduced by Upper in Ref.~\cite{Uppe97a}, and have
nearly equivalent representations sometimes known as `quasi-realizations'~\cite{Fanizza24_Quantum, Monras16_Quantum}
or `weighted automata'~\cite{Balle15_Canonical}.

To naturally connect with language models,
we are interested in
the generation and prediction of one-sided stochastic processes.  The sequence of observed tokens
($x_{1:L} = x_1 x_2 \dots x_L$ with $x_\ell \in \mathcal{X}$)
can be thought of as a realization of correlated random variables $X_{1:L} = X_1 X_2 \dots X_L \sim \Pr(X_{1:L})$,
sampled from some 
ground-truth joint distribution. 
Finite sequences can be accommodated within this framework via a special end-of-sequence token (and possibly padding tokens for after that). 
We 
focus on processes that can be generated by finite-dimensional GHMMs, since 
the question of realizability
is 
nontrivial only
when comparing the cardinality of dimensions:
Any one-sided classical stochastic process whatsoever (anywhere on the Chomsky hierarchy) can be described by an infinite-state HMM with a particular start state.

Any particular GHMM is defined by the three-tuple
$\mathcal{M} = \bigl( \mathcal{X} , \init , (T^{(x)})_{x \in \mathcal{X}} \bigr)$, 
consisting of the token alphabet $\mathcal{X}$, the initial latent vector $\init$, and a collection of linear transition operators $(T^{(x)})_{x \in \mathcal{X}}$ acting on the latent space.
The net transition operator $T = \sum_{x \in \mathcal{X}} T^{(x)}$ has an eigenvalue of unity and associated right eigenvector $\dket{1} = T \dket{1} $, and the initial row vector $\init$ is normalized such that $\dbraket{\eta^{(\varnothing)} | 1} = 1$.
To easily track object type, we use double-bras $\dbra{\cdot}$ as row vectors and double-kets $\dket{\cdot}$ as column vectors.\footnote{This notation is intentionally reminiscent of quantum Liouville-space notation, although our transposition prioritizes consistency with the Markov-chain literature.}
The probability of any sequence $w = x_{1:\ell} \in \mathcal{X}^\ell$ of arbitrary length $\ell$ 
can be 
calculated directly via linear algebra:
\begin{align}
	\Pr(X_{1:\ell} = w) = 
    \init T^{(w)} \dket{1}
    ~,
 \label{eq:Probs_from_GHMM}
\end{align}	
where $T^{(w)} = T^{(x_1)} \cdots T^{(x_\ell)} $.
(We will sometimes keep the random variable $X_{1:\ell}$ implicit for simpler notation.)
At any $\ell$, normalization in probability follows from 
$\sum_{w \in \mathcal{X}^\ell }	\Pr(w) 
=  \sum_{w \in \mathcal{X}^\ell } \init T^{(w)} \dket{1} 
=  \init T^\ell \dket{1} 
=   \dbraket{\eta^{(\varnothing)} | 1}  
= 1$.
%

Non-negativity of each word requires 
$\dbraket{\eta^{(\varnothing)} | T^{(w)} | 1}  \geq 0$
for all $w \in \mathcal{X}^*$.

\subsection{HMMs}

Hidden Markov models (HMMs) are a special case of GHMMs, where 
$(T^{(x)})_{x\in \mathcal{X}}$ are substochastic matrices 
with non-negative matrix elements that correspond to transition probabilities $T^{(x)}_{s,s'} = \Pr(s', x | s)$ from hidden state $s$ to $s'$ while observing the symbol $x$,
and these matrices then
sum to the row-stochastic transition matrix $T$.\footnote{Technically, these are Mealy HMMs, for which it is natural to think of tokens being generated during the transitions between latent states. The more common Moore HMMs, where each state has an associated probability distribution over tokens, are class-equivalent: a stochastic process with a finite Mealy HMM has a finite Moore HMM and vice versa.} 
In this case, $\dket{1}$ is then a column vector of all ones, while $\init$ is an initial probability distribution over latent states~\cite{Riec18_SSAC1}.

\subsection{QHMMs}

Repeated use of a quantum instrument yields a classical stochastic process via interaction with a quantum memory system~\cite{Monras16_Quantum, Elliott22_Quantum, Zonnios25_Quantum}.  
Typically, a quantum state is represented by a density matrix $\rho_t$ at time $t$, while linear subchannels of the transformation are often represented by Kraus operators 
$K_{x,y}$ (where $\sum_{x \in \mathcal{X}, y \in \mathcal{Y}}  K_{x,y}^\dagger K_{x,y} = I$) 
that 
yield the probability of measurement outcome $x$ (while marginalizing over possible ancillary measurement $y \in \mathcal{Y}$) via 
$\Pr(X_t =x | \rho_t) = \sum_{y \in \mathcal{Y}} \tr( K_{x,y} \rho_t K_{x,y}^\dagger)$.
Marginalizing over all possible outcomes yields a completely positive and trace preserving (CPTP) map $\rho \mapsto \sum_{x \in \mathcal{X}, y \in \mathcal{Y}} K_{x,y} \rho K_{x,y}^\dagger$ on the quantum system, whereas \emph{conditioning} on the measurement outcome $x$ yields the Bayesian update 
$\rho \mapsto \frac{\sum_{y \in \mathcal{Y}} K_{x,y} \rho K_{x,y}^\dagger }{\sum_{y \in \mathcal{Y}} \text{tr}( K_{x,y}^\dagger K_{x,y} \rho )}$.
The latter is typically nonlinear due to the normalization.


A GHMM representation of a quantum process is achieved via a linear transformation of the density matrix into a vectorized form, which also translates the effect of the Kraus operators into the $T^{(x)}$ matrices of a GHMM---either through a transposed Liouville-space representation~\cite{Gyamfi2020fundamentals}
or a generalized Bloch-vector representation~\cite{Riec24_Ideal},
as shown in App.~\ref{sec:QuantumBlochRep}.
Any classical stochastic process that can be generated with a QHMM where the quantum memory acts on a $d$-dimensional Hilbert space has a GHMM representation with no more than $d^2$ real latent dimensions~\cite{Riechers25_Identifiability}.




\section{Conditional probabilities and belief geometry}
\label{sec:ConditionalProbs}

As a consequence of Eq.~\eqref{eq:Probs_from_GHMM}, 
history-induced
conditional probability distributions over 
all possible futures $\overrightarrow{X} = X_{\ell+1: \ell+1 + L}$
for arbitrary future length $L$
are calculated as 
\begin{align}
	\Pr(\overrightarrow{X} |X_{1:\ell} = w) 
 & = 
\dbra{\eta^{(w)}}
T^{(\overrightarrow{X})} \dket{1} 
\label{eq:ConditionalLinAlg}
 ~,
\end{align}
where we define 
the \emph{predictive vector}
\begin{align}
\dbra{\eta^{(w)}} \coloneqq \frac{\init T^{(w)}}{ \init T^{(w)} \dket{1}}
\end{align}
induced by a GHMM and history $w \in \mathcal{X}^{\ell}$ that occurs with non-zero probability.\footnote{We sometimes refer to the predictive vector as a `belief state', although we emphasize that these predictive vectors do not always correspond to a probability distribution over latent variables.  Predictive vectors induced by valid histories do however always correspond to a probability density (i.e., belief) over all possible futures, via Eq.~\eqref{eq:ConditionalLinAlg}.}
%
%
This suggests a \emph{metadynamic} among predictive vectors
 $\mathcal{R} \coloneqq \Bigl\{ 
 \dbra{ \eta^{(w)}} : w \in \mathcal{X}^*, 
 \init T^{(w)} \dket{1} > 0 \Bigr\}$,
 which happens in the space orthogonal to $\dket{1}$
 (since 
 $\bigl( \dbra{\eta} - \dbra{\eta'} \bigr) \dket{1} = 0$ for all $\dbra{\eta} ,  \dbra{\eta'} \in \mathcal{R}$ ):
 From the predictive vector $\dbra{\eta^{(w)}}$, an observation $x$ induces the mapping $\dbra{\eta^{(w)}} \mapsto \dbra{\eta^{(wx)}} = \dbra{\eta^{(w)}} T^{(x)} / \dbra{\eta^{(w)}} T^{(x)} \dket{1}$.
 Notice that different histories can induce the same predictive vector, in which case they have no way to affect the future differently---these predictive vectors thus induce an equivalence class over observed histories.



Moreover, from Eq.~\eqref{eq:ConditionalLinAlg}, 
we see that \emph{when two different histories induce nearby predictive vectors, they must have a similar affect on the future}.  The geometric relationship among predictive vectors thus reflects a deeper truth about the 
linear relationship among 
\emph{process states}---the history-induced densities over futures:
\begin{align}
\dbra{\eta^{(w'')}} = c  \dbra{\eta^{(w)}} + c' \dbra{\eta^{(w')}} 
\implies
\Pr(\overrightarrow{X} | w'') = c \Pr(\overrightarrow{X} | w)  + c' \Pr(\overrightarrow{X} | w') \quad c, \, c' \in \mathbb{R} ~.
\end{align}

For an HMM, predictive states 
are points in a probability simplex (the $(|\mathcal{S}| \!-\!1)$-simplex over the hidden states of an $|\mathcal{S}|$-state HMM).
For a quantum model, 
predictive states correspond to density matrices or (via a fixed linear map from the space of density matrices) to another representation like its generalized Bloch vector or Liouville-space representation.

While $\dbra{\eta^{(w)}} = \dbra{\eta^{(w')}}$ implies that 
$\Pr(\overrightarrow{X} | w) = \Pr(\overrightarrow{X} | w') $, 
the converse is not necessarily true if the GHMM is non-minimal.
 %
Indeed,
there could still be redundancy among the predictive vectors 
of a GHMM
if 
 $\bigl( \dbra{\eta} - \dbra{\eta'} \bigr) \dket{f} = 0 $ for all $\dket{f} \in \mathcal{F} \coloneqq 
 \text{span} \bigl( \{ T^{(w)} \dket{1} \}_{w \in \mathcal{X}^*} \bigr)$
 for some $\dbra{\eta} ,  \dbra{\eta'} \in \mathcal{R}$.
 Quotienting out the relevant nullspace 
 yields the minimal set of maximally predictive features for the stochastic process.
 Viewed as points in a vector space, \emph{this minimal set of  maximally predictive features implies a fundamental geometry associated with the metadynamic of prediction.}

In the classical case of updating an HMM, the predictive vector stays in the ($|\mathcal{S}|-1$)-simplex of probability distributions over the set of hidden states $\mathcal{S}$.
Since there 
is a manifold of
pure states of a quantum system,
predictive vectors are no longer so constrained in the quantum case.  In Kraus representation, the density matrix for quantum memory updates as 
$\rho_{t+1} = \frac{\sum_y K_{x_t, y} \rho_t K_{x_t, y}^\dagger}{\sum_y \tr( K_{x_t, y} \rho_t K_{x_t, y}^\dagger)}$ given a particular measurement outcome $x_t$.  The predictive vector is simply an alternative representation of the updated quantum memory state, linearly related to the density matrix representation.
In general, the predictive vector of a GHMM is restricted to the convex hull of the pure states of the theory.  However, not all points in the convex hull are induced by observations---rather, the repeated application of Bayesian updates through each sequence fills out a (sometimes fractal) geometry of induced beliefs
in the latent memory space
of a generative model 
for the data distribution.

\section{Neural Networks Discover Minimal Belief Geometries}
\label{sec:NeuralNetsDiscoverGeometry}

We hypothesize that neural networks trained on data from a stochastic process do not merely memorize sequences, but learn an internal representation that mirrors the geometric relationships of optimal Bayesian filtering over the latent variables of minimal models of the data-generating process---whether classical, quantum, or post-quantum. To test this, we perform standard pretraining on different network architectures via next-token-prediction cross-entropy loss (for details, see Sec.~\ref{app:training-methods} of the appendix), and use a probing methodology to determine if a simple affine map exists between the network's activation vectors and the true belief-state geometry of the minimal data-generating process. Our results demonstrate that neural networks not only learn these representations, but consistently discover the most compact representation available, without any explicit knowledge of the underlying generative mechanism.

\subsection{Probing for geometric structure}

To test for learned geometric representations, we employed 
linear regression
to determine if a simple affine map exists between the network's activation vectors and the true minimal belief-state geometry of the data-generating process. For each trained network, we extracted activation vectors $\act[w] \in \mathbb{R}^{N d_\text{model}}$ by concatenating activations from all $N$ layers 
at the last token position
for context sequences $w \in \mathcal{X}^*$. We then tested whether a single affine transformation $\mathcal{L}: \mathbb{R}^{N d_\text{model}} \to \mathbb{R}^{d_\text{g}}$ could map these activations to the corresponding belief states 
$\dbra{\eta^{(w)}} \in \mathbb{R}^{d_\text{g}}$ of the minimal generator.

The optimal affine transformation was found by solving a weighted least-squares regression problem (see Appendix~\ref{app:LinRegression} for technical details). Figure~\ref{fig:main_results} presents our key finding: neural networks trained on three fundamentally different stochastic processes---requiring classical, quantum, and post-quantum minimal generators respectively---accurately learn to linearly represent their corresponding minimal belief geometries.

\subsection{Example processes with classical, quantum, and post-quantum minimal generators}

To demonstrate our main results, we focus on three example stochastic processes that require nominally classical, quantum, and post-quantum resources respectively.
 These processes and their minimal GHMMs are presented in detail in Appendix \ref{app:ExampleProcesses}.
 
\textbf{Classical}.--- The classical Mess3 process~\cite{Marz17a, Shai24_Transformers, Piotrowski25_Constrained}
 has three hidden states, three observable tokens and infinitely many belief states induced by observed histories, which arrange themselves as a fractal in the 2-simplex of probability distributions over the three latent states.

\textbf{Quantum}.---
 Our Bloch Walk process---a correlated classical stochastic process with a 4-token alphabet---can be generated with a single qubit of quantum memory but has no finite HMM representation.  The context-induced belief states live on a two-dimensional slice through the Bloch ball, and belief updates tend towards greater purity.  

 \textbf{Post-quantum}.---
The Moon process was introduced (without name) by Fanizza et al.~\cite{Fanizza24_Quantum} 
as an example of a correlated classical stochastic process 
with no finite HMM generator and no finite-dimensional quantum generator---yet it has a simple 3-dimensional GHMM generator.
 
Each of these processes are stationary and ergodic. 
For stationary processes, the initial predictive vector $\dbra{\eta^{(\varnothing)}}$ is the stationary vector $\dbra{\stationary} = \dbra{\stationary} T$, 
which is the left eigenstate of $T$ associated with the eigenvalue of unity.
Note however that our general theoretical framework accommodates both non-stationary and non-ergodic generators, both of which are relevant for language models. 

\subsection{Classical, quantum, and post-quantum belief geometries in neural network activations}

For the Mess3 process~\cite{Marz17a, Shai24_Transformers}, which admits a 3-state HMM as its minimal generator, the set of reachable belief states forms an intricate fractal structure within the 2-simplex (Fig.~\ref{fig:main_results}, top row). These belief states correspond to probability distributions over the three hidden states, with the fractal pattern emerging from the particular transition structure of the process.

Both transformer and LSTM architectures learn to represent this geometry with remarkable fidelity. The affine transformation from neural activations preserves not only the triangular boundary of the simplex but also the intricate self-similar pattern of belief states within it. The preserved color gradients---which encode relative positions in the ground-truth belief space---demonstrate that the learned representation maintains the correct geometric relationships among belief states.

Quantitatively, both architectures achieve low root mean square error (RMSE) when mapping to the true classical belief geometry (blue bars). In contrast, randomly initialized networks fail to produce this geometry (gray bars). This confirms that the geometric structure emerges specifically through training on process-generated data, rather than being an artifact of the network architecture or the high dimensionality of the activation space.


The Bloch Walk process presents a fundamental test of our hypothesis. While its minimal generator requires only a single qubit of quantum memory, any classical HMM generator would require an infinite number of states~\cite{Monras16_Quantum}. The quantum belief states for this process form a two-dimensional slice through the Bloch sphere, corresponding to density matrices with support in the $x$-$z$ plane (Fig.~\ref{fig:main_results}, middle row).

Remarkably, both transformer and LSTM networks spontaneously discover this quantum representation. Through a single affine transformation, the networks' activations map to points that accurately reproduce the circular boundary and internal structure of the Bloch disk slice. The grid pattern visible in the learned representations corresponds to the systematic exploration of the quantum state space induced by different observation sequences.

Most significantly, the networks show substantially better fit to the quantum belief geometry (blue bars) than to belief states derived from a finite-order classical Markov approximation of the process (red bars). This finite classical approximation, while capable of capturing some statistical properties of the process, requires substantially higher dimensions (we chose Markov-order-3 approximations for these analyses, since for this process the corresponding beliefs have dimensionality of the residual-stream/hidden states of the models trained) and fails to capture the true geometric relationships among belief states. The networks' preference for the quantum representation demonstrates that they are not merely learning some tractable classical approximation, but are discovering the genuinely quantum nature of the minimal representation.


For the Moon process~\cite{Fanizza24_Quantum}, even quantum generators prove insufficient---the minimal finite-dimensional generator requires a post-quantum generalized probabilistic theory (GPT). 
The belief states for this process form transient and recurrent curved manifolds that cannot arise from either classical or quantum origins (Fig.~\ref{fig:main_results}, bottom row).

Despite the exotic nature of this representation, both transformer and LSTM architectures successfully learn the post-quantum belief geometry. The networks' activations, when passed through the learned affine transformation, accurately reproduce the characteristic curved structure of the GPT belief manifold. As with the quantum case, the networks show significantly better fit to the post-quantum representation than to classical approximations, as evidenced by the lower RMSE values.

This result is particularly striking given that post-quantum theories were developed as abstract mathematical frameworks to explore the boundaries of quantum mechanics. That neural networks discover these representations naturally through gradient descent on next-token prediction suggests a deep connection between the parsimonious geometry of belief states and the computational structures that emerge in neural networks.

\subsection{Universality across architectures}

A remarkable aspect of our results is the universality of these geometric representations across fundamentally different neural architectures. Despite the stark differences between the strictly recurrent processing of LSTMs and the parallel attention mechanisms of transformers, both architectures learn essentially identical belief geometries for each process. This universality suggests that the geometric structure is not an artifact of any particular architectural inductive bias, but rather reflects the intrinsic geometry of beliefs about the data-generating process. The full results for Transformer, LSTM, vanilla RNN, and GRU architectures across all 
of our example
stochastic processes are presented in Appendix~\ref{app:extended-figs} (Figs.~\ref{fig:transformer-extended}-
\ref{fig:GRU-extended}).
Randomly initialized networks fail to produce these geometries under identical linear probing, yielding poor fits with high RMSE values (gray bars in figures). This control demonstrates that the geometric representations emerge specifically through training on next-token prediction, not from the mere expressivity of high-dimensional spaces or affine transformations.



\begin{figure}
    \centering
    \includegraphics[width=.94\linewidth]{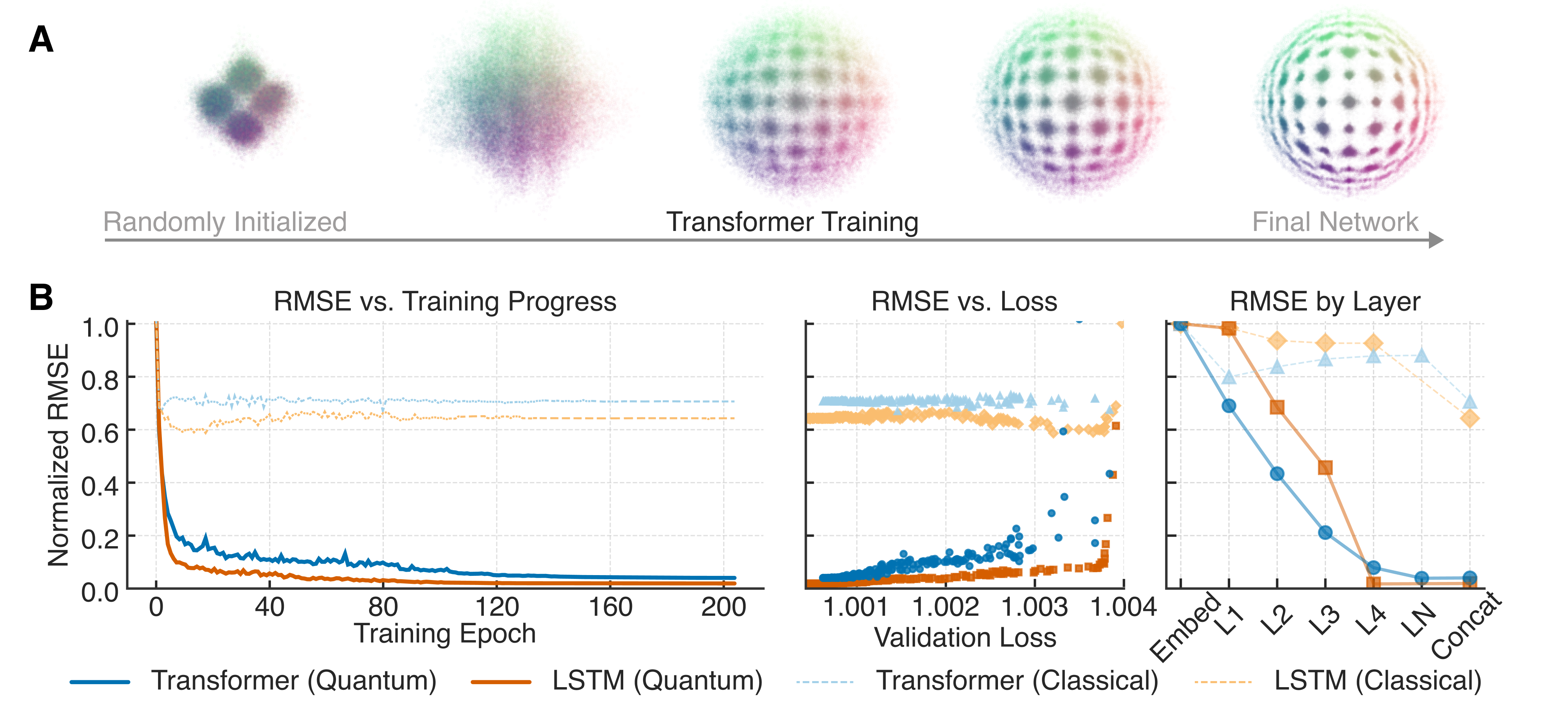}
    \caption{\textbf{Over the course of training, the geometry of activations in neural networks evolves to recapitulate the geometry of Bayesian filtering over the minimal quantum generator, and not classical approximations.} (A) Evolution of the belief geometry (linearly mapped from transformer activations) during training for the Bloch Walk process, showing convergence from a random initialization to the target 2D slice of the Bloch sphere. Colors indicate preservation of relative geometric positions. The best affine map is found anew at each epoch of training.  (B) Normalized root mean square error (RMSE, normalized to RMSE of the randomly initialized network) comparing the fit of linearly mapped activations (Transformer: blue, LSTM: orange) to the true quantum belief geometry (solid lines) versus a classical Markov-order-3 approximation (dashed lines). Plots show RMSE versus training epoch (left), validation loss (center), and network layer (right, L1-L4, LN=LayerNorm, Concat=all layers concatenated). Both architectures learn beliefs over the quantum generator well (low RMSE), especially in later layers and as validation loss decreases, while failing to represent beliefs over the classical approximation.}
\label{fig:QSlice_details}
\end{figure}

\section{Emergence of Belief Geometry During Training}

Having established that trained neural networks linearly represent minimal belief geometries, we next investigated how these representations emerge during the training process. Figure~\ref{fig:QSlice_details} tracks the evolution of geometric representations throughout pretraining, revealing that belief geometry is not an immediate consequence of network initialization but rather emerges gradually as networks learn to predict future tokens.


Panel A of Fig.~\ref{fig:QSlice_details} visualizes the evolution of belief geometry for a transformer trained on the Bloch Walk process. At random initialization, the linearly mapped activations form amorphous clouds with relatively little discernible structure. As training progresses, this cloud gradually organizes into increasingly structured patterns, ultimately converging to the characteristic 2D slice of the Bloch sphere that represents the quantum belief states of the process.

The color gradients in these visualizations encode the relative positions of belief states in the ground-truth geometry. The preservation and sharpening of these gradients during training demonstrates that the network 
is discovering the precise geometric relationships that govern how different observation histories affect predictions about the future.


Panel B of Fig.~\ref{fig:QSlice_details} provides a quantitative analysis of how geometric representations develop during training. The leftmost plot shows normalized RMSE (normalized to the RMSE of the randomly initialized network) as a function of training epoch. Both transformer and LSTM architectures exhibit a rapid decrease in RMSE during the first 20-40 epochs, with the error dropping by more than 80\% as the networks discover the quantum structure underlying the data.

Crucially, while the fit to quantum belief geometry improves dramatically (solid lines), the fit to classical Markov-order-3 approximations remains poor throughout training (dashed lines). This divergence is particularly significant: it demonstrates that networks are not simply learning better representations that happen to project onto multiple geometries, but are specifically discovering and encoding the minimal quantum representation while actively diverging from classical approximations.


The middle panel of Fig.~\ref{fig:QSlice_details}B reveals a striking correlation between the emergence of belief geometry and actual task performance. As validation loss decreases (moving left along the horizontal axis), the RMSE for quantum belief geometry systematically decreases for both architectures. This tight coupling suggests that learning to represent belief geometry is not an epiphenomenon but is intimately connected to the network's ability to predict future tokens. The relationship shows the steepest improvements in geometric representation occurring during the early phases of training when validation loss drops most rapidly. This indicates that discovering the correct belief geometry may be a key mechanism by which neural networks achieve good performance on sequence prediction tasks.


The rightmost panel examines how geometric representations vary across network layers. Both architectures show a clear progression: early layers (embedding and L1) have higher RMSE values, while later layers (L3, L4, and particularly the concatenation of all layers) achieve increasingly accurate representations of the belief geometry. This layer-wise progression suggests that belief geometry emerges through hierarchical processing. Early layers may learn local statistical patterns, while deeper layers integrate this information to construct the global geometric representation necessary for optimal prediction. 


The gradual emergence of belief geometry during training offers insights into how neural networks learn to model sequential data. Rather than simply memorizing transition statistics, networks appear to construct internal models that mirror the geometric structure of optimal Bayesian filtering. This suggests that the implicit biases of gradient descent on next-token prediction objectives naturally guide networks toward discovering minimal sufficient representations of their training data.

The fact that networks specifically learn quantum or post-quantum representations when these are the minimal generators—despite having no explicit knowledge of quantum mechanics or generalized probabilistic theories—indicates that these exotic mathematical structures may be more fundamental to sequence modeling than previously recognized. Neural networks, through the simple objective of predicting the next token, spontaneously discover the same sophisticated mathematical frameworks that physicists and mathematicians have developed to describe nature at its most fundamental level.

\section{Preservation of Geometric Relationships}

\begin{figure}
    \centering
    \includegraphics[width=0.75\linewidth]{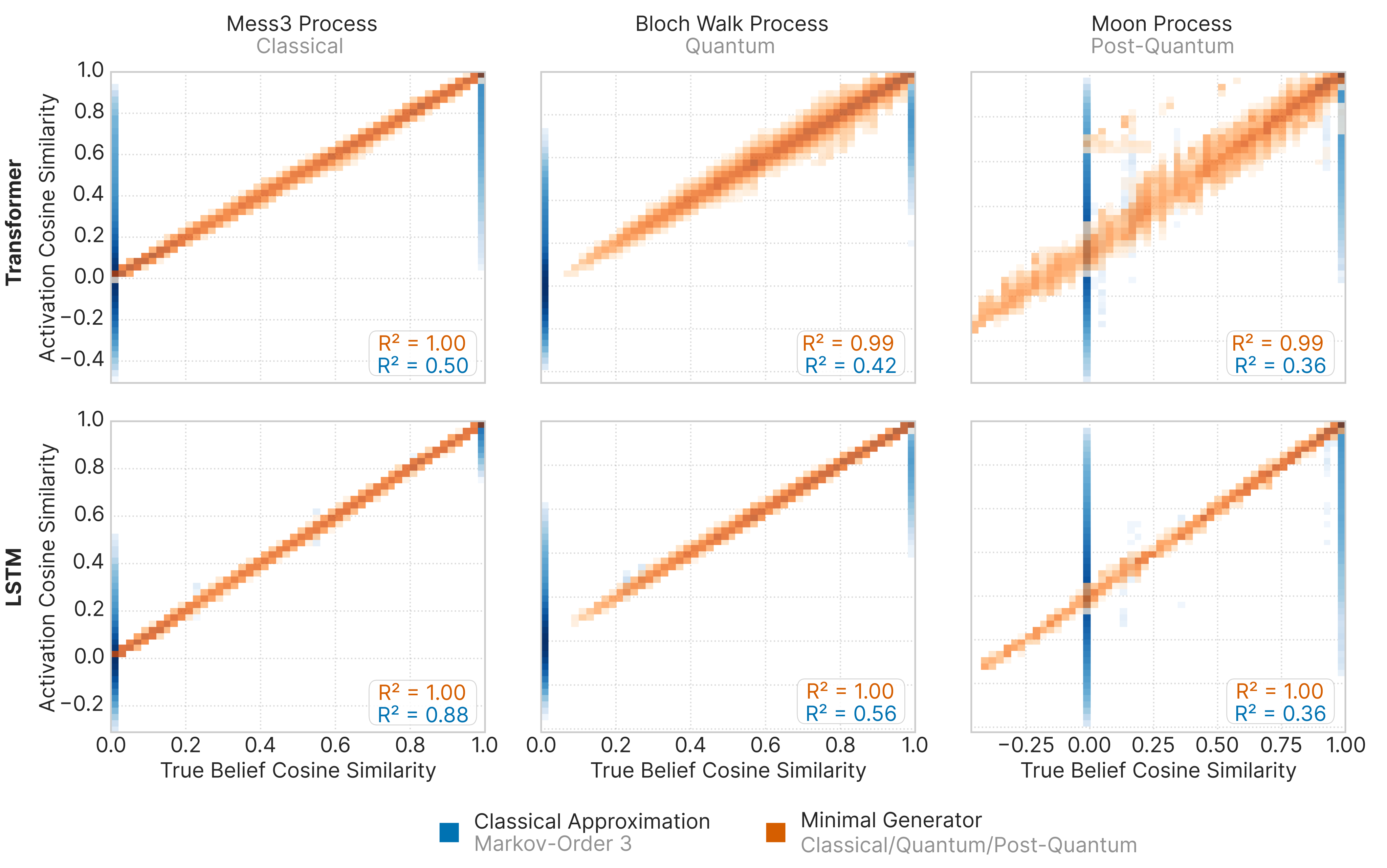}
    \caption{\textbf{Neural networks preserve geometric relationships between belief states across classical, quantum, and post-quantum processes.} Pairwise cosine similarity between belief states as represented in neural network activations (y-axis) versus the true theoretical belief geometry (x-axis). Cosine similarity measures the angular relationship between belief vectors, with 1.0 indicating parallel vectors and 0.0 indicating orthogonal vectors. Each point represents a pair of belief states induced by different context sequences, computed over all possible contexts up to the context window length. Orange points show the relationship when beliefs are taken from the minimal generator (classical for Mess3, quantum for Bloch Walk, post-quantum for Moon), while blue points show relationships for a classical Markov-order-3 approximation. Perfect preservation of geometric relationships would yield points along the diagonal. Both transformer (top row) and LSTM (bottom row) architectures show near-perfect preservation of the minimal generator geometry ($R^2 \in [0.99, 1.00]$, orange) across all three process types, while poorly representing classical approximations (blue) for the classical ($R^2 \in [0.5, 0.88]$), quantum ($R^2 \in [0.42, 0.56]$), and post-quantum ($R^2 =0.36$) processes. The tight clustering along the diagonal for minimal generators demonstrates that networks learn not just individual belief states but the complete geometric structure that governs how different observation histories relate to each other in belief space. This preservation of pairwise relationships provides strong evidence that neural networks are discovering the true underlying geometry rather than learning a distorted approximation.}
    \label{fig:RSA}
\end{figure}

Having demonstrated that neural networks learn to represent belief states and that these representations emerge during training, we now further examine the degree to which networks preserve the complete geometric structure of belief spaces. A central insight of our framework is the purposeful non-orthogonality of computationally relevant states. For quantum and post-quantum processes, the non-orthogonality of pure states 
is what enables their computational advantages over classical representations.
Non-orthogonality in these representations encode similar predictions about the future induced by different histories.


To assess whether neural networks represent these non-orthogonal states, we computed pairwise cosine similarities between all belief states in both the learned representations and the ground-truth geometry. Figure~\ref{fig:RSA} shows these comparisons for all three processes across both architectures. Each point represents a pair of contexts, with the x-coordinate showing their cosine similarity in the true belief geometry and the y-coordinate showing their similarity after the learned affine mapping from neural activations.

For networks trained on 
token sequences
generated from classical, quantum, and post-quantum processes, the cosine similarities cluster tightly along the diagonal with $R^2$ values between 0.99 and 1.00. This near-perfect correlation demonstrates that the learned affine mapping preserves not just individual belief states but the entire web of relationships among them. The networks have discovered 
representations 
that preserve angles and relative distances 
among ground-truth predictive vectors
with remarkable fidelity.

In stark contrast, when we attempt to map network activations to 
Bayesian-updated belief states in the simplex of
classical Markov approximations of the quantum and post-quantum processes (blue points), the geometric preservation fails dramatically. The $R^2$ values drop to between 0.36 and 0.56, indicating that the learned representations are fundamentally incompatible with classical belief structures. The vertical bands at 0 and 1 in the classical approximation plots are due to the fact that in finite-order Markov models of these processes, many belief states are either identical or orthogonal. This discrete structure 
of the classical approximations
cannot capture the continuous manifold of belief states that characterizes the true underlying geometry implied by quantum and post-quantum computation.

This geometric analysis reveals that neural networks do not simply memorize sequences without relational understanding.
Rather, they learn self-consistent geometric embeddings of how every context relates to every other.
The high fidelity with which networks preserve geometric relationships suggests they have discovered how to represent and transform between belief states in ways that respect the non-orthogonal structure of minimal generators—even when these correspond to quantum measurements or post-quantum operations that have no finite classical implementation.

The consistency of these results across both transformer and LSTM architectures reinforces our findings about universality. Despite their different computational mechanisms, both architectures discover representations that preserve the same geometric relationships. The near-perfect preservation of geometric structure indicates that this is not a loose approximation but a precise discovery—the networks have found representations whose internal geometry mirrors the theoretical optimal solution with extraordinary accuracy.

We note that this 
learned belief
geometry reflects the Platonic form of the statistical structure of the training data, 
imprinted in the activations of pretrained neural networks regardless of the particulars of their architecture.
This discovered universal representation
supports and refines the 
Platonic representation hypothesis~\cite{Huh24_Position}.

\section{Discussion}

\subsection{Universal geometric representation---Why?}

Many different models---classical, quantum, or post-quantum, designed or evolved---can produce exactly the same stochastic process~\cite{Crut10a}.
However,
the geometric structure of minimal belief states is 
independent of the 
generative model
and, rather, only depends on the \emph{process} generated by the model. 
Indeed, the neural network never sees the choice of generator representation---it only sees data.  So how is it that this canonical geometric representation emerges in the activations of deep neural networks?
The answer lies in the linear relationships among conditional probability densities themselves.
Each probability density over the entire set of possible futures can be thought of as a point in an abstract vector space.
Observations and Bayesian updates induce transformations among these vectors.

It is notable too that the canonical geometric representation of beliefs about the future is universal across neural network architectures, as we have shown empirically.  This universality was first predicted in Ref.~\cite{Shai24_Transformers},  confirmed for classical processes in the case of both transformers \cite{Shai24_Transformers} and subsequently RNNs \cite{Pepper24_RNNs},
and now extended to RNNs, LSTMs, and transformers, across a wide range of hyperparameters, for a wide range of processes---including those processes that only have finite-dimensional representations in nominally quantum and post-quantum theories.
It is remarkable that the belief geometry is invariant 
despite
the stark differences among strictly-recurrent, strictly feed-forward, and hybrid architectures.

\subsection{Refining relative advantages of quantum and classical}



In the preceding, we have shown that, for all practical purposes (i.e., up to floating-point machine precision), neural nets learn quantum and post-quantum representations---and, importantly, 
their activations linearly represent the 
geometry of beliefs 
that would be attained from using these 
quantum and post-quantum models
to update beliefs about the future as they move through the sequence of classical tokens.
%
Since we have shown that neural networks can implement some analog of quantum processing, this calls for a clarification of where genuine quantum processing (i.e., with real qubits and a quantum computer) provides computational advantages over neural networks.
Do our results 
suggest that classical neural networks can be efficient replacements for future quantum computers?  No.


The true practical advantage of genuinely quantum systems will be the exponential growth of the dimension of the Hilbert space dimension as quantum subsystems are composed.  
Recall that neural nets need no more than $d^2$ dimensions to represent a quantum model with a Hilbert space of dimension $d$.  E.g., if a quantum model requires $N$ qubits, then the Hilbert space dimension is $2^N$ and a corresponding GHMM may need up to $2^{2N}$ dimensions (although there will sometimes be smaller post-quantum representations).
This scaling may be disappointing for those who would like to 
draw inspiration from our work to
use classical neural nets in place of large quantum circuits.
However, this scaling is a relief
in the sense that RSA encryption (the basis of much modern cryptography) will not be immediately broken by a classical neural network imitating a quantum circuit---since it is estimated that at least 4000 logical qubits would be required for quantum prime-factoring algorithms, which suggests a neural network with at least $2^{4001} - 2$ neurons would be required to implement the same circuit,\footnote{Why $2^{4001} - 2$ neurons
rather than $2^{8000}$?  Indeed, our results imply that there exists a GHMM with $2^{2N} = 2^{8000}$ real latent dimensions that can replicate a process with a quantum memory that acts on a $N=4000$ dimensional Hilbert space.  But since the quantum state of a standard idealized quantum circuit is always pure, even after going through many quantum gates, the number of real coefficients needed to specify the quantum state is $2\times 2^N$, since there are $d = 2^N$ complex coefficients needed to specify an arbitrary pure quantum state of $N$ qubits.  After subtracting one dimension for the normalization constraint and subtracting one dimension for global phase invariance, we arrive at our overly-precise estimate of $2^{N+1} - 2 = 2^{4001} - 2$ real parameters that would seem to lower bound the number of neurons necessary in a neural network that implements the desired quantum circuit.  Again, with $2^{8000}$ dimensions, there is a mathematically constructible GHMM and a corresponding representation that can be embedded within a neural network with as many dimensions---but we don't even have enough atoms in the universe to construct a neural network with 
the relatively meager number of $2^{4001}$ neurons.  Breaking modern RSA encryption will need to wait for quantum computers with error correction.} which is infeasible due to an insufficient number of atoms in the observable universe (if each classical neuron requires at least one atom).

However, there remains significant nuance in what constitutes a genuinely quantum advantage.
For instance, we have shown that small neural networks can effectively (i.e., up to machine precision) represent a post-quantum model, and its implied belief geometry, with only few neurons---even though, if we required exact precision, there is no finite-dimensional classical nor quantum model up to the task.
One possible 
takeaway for the scientific community at large is 
that 
claims of 
a quantum advantage should carefully 
describe whether and to what extent it implies a practical advantage over classical computing systems that, via floating point numbers, have access to an effectively real-valued vector space.





\section{Conclusion}

Among the networks we train are generative pretrained transformers (also known as `GPTs'), 
yielding the cute lesson that ``GPTs learn GPTs''---i.e., \emph{generative pretrained transformers (GPTs) learn (and represent belief states over) generalized probabilistic theories (GPTs)
that efficiently represent their training data}.
But the same is true for other neural-network architectures too, as long as they are pretrained as usual on 
next token prediction.

It turns out that neural networks' access to an effectively real-valued vector space gives them abilities outside the scope
of naively-defined 
`classical' models with discrete memory elements.
%
This significantly updates our understanding of the nature of computation in neural networks:
Not only do pretrained neural networks effectively perform Bayesian updates over the latent states of a world model as they observe more context during inference---but, moreover, their world models may leverage quantum representations, even when the training data is fully classical. 

Our 
results about post-quantum representations 
update the discussion of neural networks' ability to learn world models---both for proponents and opponents of such world-model claims.  
For some classical stochastic processes, 
we see that neural nets do indeed learn a form of world model, but not one that behaves according to the physics of a classical or even quantum reality.

As more tokens are observed, 
the neural activation pattern reflects updated beliefs about the entire future 
as the sequence model homes in 
on the latent state of 
the world 
through increasing context.
Independent of neural-network architecture, we find that context-induced neural activation vectors 
relate to each other according to 
universal 
belief geometries---high-dimensional arrangements of non-orthogonal patterns reflecting Bayesian-updated beliefs over the latent dimensions of classical, quantum, or post-quantum models capable of generating the training data.
Looking ahead we note that, despite this universality, the way that specific architectures implement this effective Bayesian updating likely affects how they generalize out of distribution~\cite{Piotrowski25_Constrained, Riechers25_Next}.

\section*{Acknowledgments}

The authors benefited from discussions with many wonderful colleagues.
TJE is supported by the University of Manchester Dame Kathleen Ollerenshaw Fellowship.

\appendix

\section{Classical, quantum, and post-quantum states:  Beliefs beyond the simplex}

While the belief states of a 
classical model live in a probability simplex (a type of polytope), belief states over quantum and post-quantum states can live in more general cross sections of generalized cones that can't be described by a finite number of vertices.  The contrast between classical and quantum memory states offers an easy analogy.  A classical bit has only two pure states---zero and one.  If we have some uncertainty about its state, then it can be in a probabilistic mixture of these two states---a convex combination that lives on the one-simplex.
In contrast, a quantum bit (i.e., a `qubit') has an uncountably infinite number of distinct pure states $\{ \ket{\psi} \! \bra{\psi} :  \ket{\psi} = c_0 \ket{0} + c_1 \ket{1}; \, |c_0|^2 + |c_1|^2 = 1; \, c_0, c_1 \in \mathbb{C} ; \, \bra{\psi} = \ket{\psi}^\dagger \}$, typically represented as the surface of the Bloch sphere---Note that while $\ket{\psi}$ is a linear combinations of $\ket{0}$ and $\ket{1}$, the state $\ket{\psi} \! \bra{\psi}$ as a rank-1 operator of trace 1 is nevertheless not a convex combination of any other pure states (which are all restricted to rank-1 operators of trace 1).  Meanwhile the interior of the Bloch ball corresponds to all possible distinct non-pure mixed states of a qubit---convex combinations of pure states (although any non-pure quantum mixed state has infinitely many different convex combinations that lead to it).
More generally, pure states are the 
extremal states that 
cannot be obtained by convex combinations of other states.  
%

\section{Quantum Representations: From density matrices and Kraus operators to GHMMs via either Liouville-space or generalized Bloch representations}
\label{sec:QuantumBlochRep}

\subsection{(Transposed) Liouville-space representation}

We obtain the transpose of the standard Liouville-space representation~\cite{Gyamfi2020fundamentals} 
via the linear invertible ket-flipper map $\mho$,
such that $\mho(c \ket{\alpha} \! \bra{\beta}) = c \ket{\alpha}^\top \otimes \bra{\beta} = \bra{\alpha}^* \! \otimes c \bra{\beta} $ 
for all $c \in \mathbb{C}$ and for any two vectors in the Hilbert space $\ket{\alpha}, \ket{\beta} \in \mathcal{H}$, where $\bra{\alpha} = \ket{\alpha}^\dagger$ and $(\cdot)^*$ denotes complex conjugation.
In this case, our GHMM is constructed from the initial vector $\init = \mho(\rho_0)$
and from the transition operators
$T^{(x)} = \sum_y K_{x,y}^\top \otimes K_{x,y}^\dagger$.
The resulting net transition operator 
$T = \sum_{x \in \Abet} T^{(x)}$
has the stationary right 
eigenstate $\dket{1} = \sum_{\zeta} \ket{\zeta}^* \otimes \ket{\zeta}$
for any orthonormal basis $\{ \ket{\zeta}\}_\zeta$ such that $\braket{\zeta'| \zeta} = \delta_{\zeta', \zeta}$, corresponding to the standard trace functional of quantum mechanics.


\subsection{Generalized Bloch representation of density matrices}

It is well known that the state of a qubit $\rho$ can be expressed via its Bloch vector $\vec{a}$:
\begin{align}
	\rho = I/2 + \vec{a} \cdot \vec{\sigma} / 2 ~,
	\label{eq:BlochQubit}
\end{align}
where 
$\vec{\sigma} = (\sigma_x, \sigma_y, \sigma_z)$ is the vector of Pauli matrices.	
For a quantum system of arbitrary finite dimension---i.e., a qudit $\rho$ acting on
a $d$-dimensional vector space $\mathcal{V}_d$---we 
achieve something similar via a 
generalized Bloch decomposition
\cite{Jako01, Riec24_Ideal}. 
We choose any complete basis 
$(I/d, \Gamma_1, \Gamma_2, \dots \Gamma_{d^2 - 1})$ for linear operators acting on $\mathcal{V}_d$,
such that the Hermitian operators $\Gamma_n$ are all traceless and mutually orthogonal, satisfying
\begin{subequations}
	\begin{align}
		\tr( \Gamma_n) &= 0 , \; \text{ and } 
		\label{eq:traceless}  \\	
		\tr(\Gamma_m \Gamma_n) &=  \constname \, \delta_{m,n} ~,
		\label{eq:orthonormal}
	\end{align}
\end{subequations}
where we will choose the normalizing constant to be $\constname = \tfrac{d-1}{d}$.
Any density matrix then has a unique decomposition in the 
operator basis 
$\vec{\Gamma} = (\Gamma_1, \Gamma_2, \dots \Gamma_{d^2 - 1})$, 
described by the 
\emph{generalized Bloch vector} $\vec{b} \in \mathbb{R}^{d^2 - 1}$
via 
\begin{align}
	\rho 
	&= \sysref + \vec{b} \cdot \vec{\Gamma} \\
	&= \Bigl( 
	\bigl[ 1 \; \vec{b} \bigr] 
	\otimes I \Bigr) 
	\begin{bmatrix}
	I/d \\
	\Gamma_1 \\
	\vdots \\
	\Gamma_{d^2 - 1}	
	\end{bmatrix}	
	~,
	\label{eq:GenBlochExpansion}
\end{align}
where each ``$I$'' above is the $d$-dimensional identity matrix.
Density matrices are thus 
a linear function of 
their 
$d^2$-dimensional
\emph{extended Bloch vector}
$\bigl[ 1 \; \vec{b} \bigr] $.
This linear function is determined uniquely from the choice of operator basis.
Conversely, given any Hermitian operator $M = cI/d + \vec{b} \cdot \vec{\Gamma}$,
its extended Bloch vector $\bigl[ c \; \vec{b} \bigr]$ 
can be obtained via
\begin{align}
	c = \tr(M) \quad \text{and } \quad \vec{b} = \tr(M \vec{\Gamma}) / \constname ~.
\end{align}	

Since the magnitude of the
Bloch vector is 
$b = \sqrt{\frac{\tr(\rho^2) d - 1}{d-1}}$,
the density matrix represents a pure state iff
the magnitude of its corresponding Bloch vector is one.  For $d>2$, not all points in the Bloch ball correspond to physical states, but the set of all physical states is nevertheless a convex set---the convex hull of the pure states, which all lie on a $2(d-1)$-dimensional submanifold of the $(d^{2}-2)$-dimensional surface of the Bloch sphere~\cite{Jako01}.

We recover the standard Bloch vector representation
in the familiar two-dimensional case of a qubit
when 
$\vec{\Gamma} = \vec{\sigma}/2 = ( \sigma_x/2, \, \sigma_y/2, \, \sigma_z/2 )$ and $\constname = 1/2$.



\subsection{Generalized Bloch representation of subchannels}

If we marginalize over measurements, the memory goes through
a completely positive and trace preserving (CPTP) map.  However, because of the measurement, the CPTP map has 
$|\mathcal{X}|$ subchannels, each a trace-non-preserving superoperator $A_x$ on the density matrix.
By linearity, each of these
subchannels 
can be fully determined by
measuring the output from $d^2$ independent inputs to the subchannel.  
Moreover, each subchannel has a generalized Bloch representation that can be readily constructed from such experiments.

Given access to the subchannels $(A_x)_{x \in \mathcal{X}}$ (either directly or via post-selection), and $d^2$ linearly independent 
input memory states $(\rho_{(n)})_{n=1}^{d^2}$,
we can build up the Bloch representation of the process itself.
Notice that each subchannel $A_x$ is a linear operator on
a density matrix, which is in turn a linear function of its extended Bloch vector.  Accordingly, each subchannel can just as well be represented as a linear operator $G^{(x)}$ acting on the extended Bloch vector:
\begin{align}
\bigl[ 1 \; \vec{b}_n \bigr] G^{(x)} = \bigl[ c_n \; \vec{b}_n' \bigr] ~,
\end{align}	
where $\bigl[ 1 \; \vec{b}_n \bigr]$ is the extended Bloch vector of $\rho_{(n)}$, and $ \bigl[ c_n \; \vec{b}_n' \bigr] $ is the extended Bloch vector of $A_x(\rho_{(n)}) = \sum_{y} K_{x,y} \rho_{(n)} K_{x,y}^\dagger$, 
with $c_n = \tr \bigl[ A_x(\rho_{(n)}) \bigr]$
and 
$\vec{b}_n' =  \tr \bigl[ A_x(\rho_{(n)}) \vec{\Gamma} \bigr] / \constname$.

Stacking these equations for the $d^2$ linearly independent memory inputs
\begin{align}
	\underbrace{
	\begin{bmatrix}
		1 & \vec{b}_1 \\
		1 & \vec{b}_2 \\
		\vdots & \vdots \\
		1 & \vec{b}_{d^2} 
	\end{bmatrix}	
}_{\eqqcolon B}
G^{(x)}
&= 
	\underbrace{
	\begin{bmatrix}
		c_1 & \vec{b}_1' \\
		c_2 & \vec{b}_2' \\
		\vdots & \vdots \\
		c_{d^2} & \vec{b}_{d^2}' 
	\end{bmatrix}	
}_{\eqqcolon B'}
 ~,
\end{align}	
and recording the new extended Bloch matrices $B$ and $B'$,
allows us to directly construct $G^{(x)}$: 
\begin{align}
G^{(x)} = B^{-1} B' ~.
\end{align}	
Notice that $B$ is always invertible, 
since we have insisted on $d^2$ linearly independent input states (which is a generic outcome of selecting $d^2$ such matrices at random).

This construction directly yields a $d^2$-dimensional GHMM-like representation of the stochastic process.
We obtain the net transition operator $G = \sum_{x \in \mathcal{X}} G^{(x)}$,
and the stationary vectors of the process $\dbra{\stationary} = \dbra{\stationary} G = \bigl[ 1 \;  \vec{b}_{\stationary} \bigr]$
and $\dket{1} = G \dket{1} = \bigl[ 1 \; 0 \, \dots \, 0 \bigr]^\top$.

Just as in Eq.~\eqref{eq:Probs_from_GHMM},
stationary probabilities are obtained via
\begin{align}
	\Pr(X_{1:L} = x_{1:L}) = \dbra{\stationary} G^{(x_{1:L})} \dket{1} ~,
\end{align}	
with 
$G^{(x_{1:L})} = G^{(x_{1})} \cdots G^{(x_{L})} $.

\section{Shared properties across quantum representations}


How unique are the generalized Bloch representations of a process?
Indeed, other representations can be easily obtained by changing the Hermitian operator basis.
Belief geometry associated with minimal generative models is nevertheless unique up to linear transformation of the space.

More generally, different representations will all have the same spectral properties on their non-zero eigenspaces, so long as those eigenspaces are not in the nullspace of the history and future spaces.
The belief geometry will be unique up to a linear transformation---and so the geometric relationship among beliefs will be qualitatively preserved across representations.

When a single Kraus operator is associated with an observation $x$---i.e., when $\sum_y K_{x,y} \otimes K_{x,y}^* = K_x \otimes K_x^*$---then the spectral properties of the transition operator $T^{(x)}$ are inherited from the spectral properties of the Kraus operator $K_x$.
From the properties of the tensor product, it immediately follows from the Liouville-space representation that the eigenvalues of the 
transition operators $T^{(x)}$ are all the possible products of the eigenvalues $\Lambda_{K_x}$ of the corresponding Kraus operator and its complex conjugates: $\Lambda_{T^{(x)}} = \bigcup_{\lambda , \zeta \in \Lambda_{K_x}} \{ \lambda^* \zeta \}$.
Moreover, the eigenvectors of $T^{(x)}$ can similarly be composed from the tensor products of the $K_x$ left and right eigenvectors with other complex-conjugated left and right eigenvectors of $K_x$.



 \section{Example processes}
\label{app:ExampleProcesses}

\subsection{Classical: Mess3 process}
 
The Mess3 process~\cite{Marz17a, Shai24_Transformers, Piotrowski25_Constrained} has three hidden states 
 $\SSet = \{ 1, 2, 3\}$, and
three observable tokens $\mathcal{X} = \{ a, b, c \}$.

The process is defined by two parameters,
 $\alpha$ and $x$,
 with dependent quantities
 $\beta = (1-\alpha)/2$ and $y = 1-2x$. For experiments we used $x=0.05, \alpha=0.85$.

 The labeled transition matrices are:
 \begin{align}
 T^{(a)} &= 
 \begin{bmatrix}
 	\alpha y & \beta x & \beta x \\
 	\alpha x & \beta y & \beta x \\
 	\alpha x & \beta x & \beta y
 \end{bmatrix}
 \\
  T^{(b)} &= 
 \begin{bmatrix}
 	\beta y & \alpha x & \beta x \\
 	\beta x & \alpha y & \beta x \\
 	\beta x & \alpha x & \beta y
 \end{bmatrix}
 \\
   T^{(c)} &= 
 \begin{bmatrix}
 	\beta y & \beta x & \alpha x \\
 	\beta x & \beta y & \alpha x \\
 	\beta x & \beta x & \alpha y
 \end{bmatrix} ~.
 \end{align}

\subsection{Quantum: Bloch Walk process}





Here we introduce the Bloch Walk process, 
a probability density over sequences of tokens $\mathcal{X} = \{ 0,1,2,3\}$,
which can be generated with a single qubit of quantum memory but has no finite HMM representation.

We find that neural networks trained on this classical stochastic process
(whether RNNs or transformers)
linearly represent the predicted 
Bloch vector of the qubit 
state of the minimal quantum memory 
capable of producing this process.
As anticipated, the 
fractal belief geometry in the Bloch sphere is linearly represented in the activation of the neural networks.

Any $y$-component of the Bloch vector has no affect on the dynamics, and any such $y$-component initially present in quantum memory decays exponentially.
However, for the stationary process we train on, the optimal belief states always live in the $x$-$z$ slice of the Bloch ball.

Note that the belief geometry over the minimal quantum representation of this process
lives in a two-dimensional subspace of the Bloch sphere, whereas the minimal classical-computational representation of this process would require an infinite number of dimensions.

\subsubsection{Kraus operators}

The process is parametrized by $\alpha > 0$ and $\beta \in \mathbb{R}$.
Let $\gamma = 1/(2 \sqrt{\alpha^2 + \beta^2} )$
We will take $\alpha = 1$ and $\beta > 1$. For experiments we used $\alpha=1, \, \beta=\sqrt{51}$.

The classical stochastic process is fully described via the following four observable-indexed Kraus operators that act on the quantum memory:
\begin{align}
K_0 &= \gamma \begin{bmatrix}
    \alpha+\beta & 0 \\
    0 & \alpha - \beta
\end{bmatrix}
= 2 \gamma \alpha (I/2) + 2 \gamma \beta (\sigma_z / 2) \\
K_1 &= \gamma \begin{bmatrix}
    \alpha-\beta & 0 \\
    0 & \alpha + \beta
\end{bmatrix}
= 2 \gamma \alpha (I/2) - 2 \gamma \beta (\sigma_z / 2) \\
K_2 &= \gamma \begin{bmatrix}
    \alpha & \beta \\
    \beta & \alpha 
\end{bmatrix}
\qquad \quad \; \, = 2 \gamma \alpha (I/2) + 2 \gamma \beta (\sigma_x / 2) \\
K_3 &= \gamma \begin{bmatrix}
    \alpha & -\beta \\
    -\beta & \alpha 
\end{bmatrix}
\qquad = 2 \gamma \alpha (I/2) - 2 \gamma \beta (\sigma_x / 2) 
~.
\end{align}

These Kraus operators satisfy
$\sum_{n=0}^3 K_n^\dagger K_n = I$.
Each Kraus operator induces a non-trace-preserving superoperator on the quantum memory state $A_n(\rho) = K_n \rho K_n^\dagger$.
In this case, since $\alpha$ and $\beta$ are real, $K_n^\dagger = K_n$.
We also note the linear dependence
$K_3 = K_0 + K_1 - K_2$.

From the initial fully mixed state, the belief would move towards $+\ket{z}$,
$-\ket{z}$,
$+\ket{x}$,
or 
$-\ket{x}$,
upon seeing the token
0, 1, 2, or 3, respectively.
For this process, 
the density matrix for quantum memory updates as 
$\rho_{t+1} = \frac{K_{n_t} \rho_t K_{n_t}^\dagger}{\tr( K_{n_t} \rho_t K_{n_t}^\dagger)}$ given a particular measurement outcome $n_t \in \mathcal{X}$.

\subsubsection{Bloch representation}

As shown in Sec.~\ref{sec:QuantumBlochRep},
we can find the GHMM-like extended Bloch representation of this process 
via the extended Bloch vectors of four linearly independent memory inputs to each subchannel, as well as the extended Bloch vectors of the subseqeunt outputs.

For our qubit memory, we can represent its state in the Hermitian operator basis $(I/2, \vec{\sigma}/2) = (I/2, \sigma_x/2, \sigma_y/2, \sigma_z/2)$, with the standard Pauli matrices
\begin{align}
\sigma_x &= 
\begin{bmatrix}
    0 & 1 \\
    1 & 0 
\end{bmatrix} ~, \\
\sigma_y &= 
\begin{bmatrix}
    0 & -i \\
    i & 0 
\end{bmatrix} ~, \text{ and } \\
\sigma_z &= 
\begin{bmatrix}
    1 & 0 \\
    0 & -1 
\end{bmatrix} ~.
\end{align}

In fact, we can input the operator basis itself in this case since we can calculate everything analytically. 
This yields the convenient Bloch matrix $B = I$ for the input operators:

\begin{align}
	\underbrace{
	\begin{bmatrix}
		c_1 & \vec{b}_1 \\
		c_2 & \vec{b}_2 \\
		\vdots & \vdots \\
		c_4 & \vec{b}_{4} 
	\end{bmatrix}	
}_{\eqqcolon B}
G^{(n)}
=
G^{(n)}
&= 
	\underbrace{
	\begin{bmatrix}
		c_1' & \vec{b}_1' \\
		c_2' & \vec{b}_2' \\
		\vdots & \vdots \\
		c_{4}' & \vec{b}_{4}' 
	\end{bmatrix}	
}_{\eqqcolon B'}
 ~.
 \label{eq:TomProcessBlochMatrices}
\end{align}	

Each output row 
on the right-hand side of Eq.~\eqref{eq:TomProcessBlochMatrices}
is the extended Bloch representation of $A_n(I/2)$ (for the first row) or $A_n(\vec{\sigma})$ (for the last three rows).
In particular,
$c_m' = \tr \bigl[ A_n(\rho_{(m)}) \bigr]$
and 
$\vec{b}_m' =  \tr \bigl[ A_n(\rho_{(m)}) \vec{\sigma} \bigr] $.
To calculate these, it will be useful to recall 
the relevant Cayley table:
\begin{align}
\begin{bmatrix}
    \sigma_x \\
    \sigma_y \\
    \sigma_z
\end{bmatrix}
\begin{bmatrix}
    \sigma_x &
    \sigma_y &
    \sigma_z
\end{bmatrix}
&= 
\begin{bmatrix}
    I & i \sigma_z & -i \sigma_y \\
    -i \sigma_z & I & i \sigma_x \\
    i \sigma_y & -i \sigma_x & I 
\end{bmatrix} ~.
\end{align}

Following through the algebra,
$A_n(\rho_{(m)}) = K_n \rho_{(m)} K_n^\dagger = \gamma^2 (\alpha I \pm \beta \sigma_{x/z}) \rho_{(m)} (\alpha I \pm \beta \sigma_{x/z})$,
where $\rho_{(m)} \in (I/2, \vec{\sigma}/2) $.

We find 
%
\begin{align}
G^{(0)} &= 
\begin{bmatrix}
    1/4 & 0 & 0 & \phantom{-}2\alpha \beta \gamma^2 \\
    0 & (\alpha^2 - \beta^2)\gamma^2 & 0 & 0 \\
    0 & 0 & (\alpha^2 - \beta^2)\gamma^2 & 0 \\
    \phantom{-} 2\alpha \beta \gamma^2 & 0 & 0 & 1/4 
\end{bmatrix} \\
G^{(1)} &= 
\begin{bmatrix}
    1/4 & 0 & 0 & -2\alpha \beta \gamma^2 \\
    0 & (\alpha^2 - \beta^2)\gamma^2  & 0 & 0 \\
    0 & 0 & (\alpha^2 - \beta^2)\gamma^2  & 0 \\
    -2\alpha \beta \gamma^2 & 0 & 0 & 1/4 \\
\end{bmatrix} \\
G^{(2)} &= 
\begin{bmatrix}
    1/4 & \phantom{-}2\alpha \beta \gamma^2 & 0 & 0 \\
    \phantom{-}2\alpha \beta \gamma^2 & 1/4 & 0 & 0 \\
    0 & 0 & (\alpha^2 - \beta^2)\gamma^2  & 0 \\
    0 & 0 & 0 & (\alpha^2 - \beta^2)\gamma^2 
\end{bmatrix} ~, \text{ and } \\
G^{(3)} 
&= 
\begin{bmatrix}
    1/4 & -2\alpha \beta \gamma^2 & 0 & 0 \\
    -2\alpha \beta \gamma^2 & 1/4 & 0 & 0 \\
    0 & 0 & (\alpha^2 - \beta^2)\gamma^2 & 0 \\
    0 & 0 & 0 & (\alpha^2 - \beta^2)\gamma^2  
\end{bmatrix} ~.
\end{align}
(Note that $G^{(3)} \neq G^{(0)} + G^{(1)} - G^{(2)} $, even though $K_3 = K_0 + K_1 - K_2$.)

The net transition operator 
\begin{align}
G = \sum_{n \in \mathcal{X}} G^{(n)}
&= 
\begin{bmatrix}
    1 & 0 & 0 & 0 \\
    0 & 1/2 + 2 (\alpha^2 - \beta^2)\gamma^2 & 0 & 0 \\
    0 & 0 & 4(\alpha^2 - \beta^2)\gamma^2  & 0 \\
    0 & 0 & 0 & 1/2 + 2(\alpha^2 - \beta^2)\gamma^2 
\end{bmatrix}
\end{align}
has the stationary extended Bloch vector $\dbra{\stationary} = \dbra{\stationary} G = \begin{bmatrix} 1 & 0 & 0 & 0 \end{bmatrix}$ associated with the fully mixed quantum state $I/2$. 
Starting from this initial belief over the quantum memory, the Bayesian-updated extended Bloch vector (upon sequential observations)
remains in the subspace spanned by $\{ I, \sigma_x, \sigma_z \}$.
Accordingly the belief metadynamics of the stationary stochastic process exists in a three-dimensional linear subspace and moreover, due to the normalization of quantum-state density matrices, in only a two-dimensional affine subspace corresponding to the $x$-$z$ plane of the Bloch sphere.


The minimal three-dimensional GHMM for this process is thus easily obtained by projecting out the non-utilized $y$-component of the Bloch sphere:
%
\begin{align}
T^{(0)} &= 
\begin{bmatrix}
    1/4 & 0 & \phantom{-}2\alpha \beta \gamma^2 \\
    0 & (\alpha^2 - \beta^2)\gamma^2  & 0 \\
    \phantom{-} 2\alpha \beta \gamma^2 & 0 & 1/4 
\end{bmatrix} \\
T^{(1)} &= 
\begin{bmatrix}
    1/4 & 0 & -2\alpha \beta \gamma^2 \\
    0 & (\alpha^2 - \beta^2)\gamma^2  & 0 \\
    -2\alpha \beta \gamma^2 & 0 & 1/4 \\
\end{bmatrix} \\
T^{(2)} &= 
\begin{bmatrix}
    1/4 & \phantom{-}2\alpha \beta \gamma^2 &  0 \\
    \phantom{-}2\alpha \beta \gamma^2 & 1/4 &  0 \\
    0 & 0 & (\alpha^2 - \beta^2)\gamma^2 
\end{bmatrix} ~, \text{ and } \\
T^{(3)} 
&= 
\begin{bmatrix}
    1/4 & -2\alpha \beta \gamma^2 & 0 \\
    -2\alpha \beta \gamma^2 & 1/4 & 0 \\
    0 & 0 & (\alpha^2 - \beta^2)\gamma^2  
\end{bmatrix} ~,
\end{align}
which can be interpreted as acting from the right on the coefficients $[c,b_x, b_z]$ of the ordered operator basis $(I/2, \, \sigma_x/2, \, \sigma_z/2)$.

The net transition operator 
\begin{align}
T = \sum_{n \in \mathcal{X}} T^{(n)}
&= 
\begin{bmatrix}
    1 & 0 & 0 \\
    0 & 1/2 + 2 (\alpha^2 - \beta^2)\gamma^2 & 0 \\
    0 & 0 & 1/2 + 2(\alpha^2 - \beta^2)\gamma^2 
\end{bmatrix}
\end{align}
has the stationary vectors $\dbra{\stationary} = \dbra{\stationary} T = \begin{bmatrix} 1 & 0 & 0 \end{bmatrix}$
and $\dket{1} = T \dket{1} = \begin{bmatrix}
    1 & 0 & 0
\end{bmatrix}^\top$.


\subsection{Quantum: FRDN}


%
%
%

Despite not having any finite HMM generator,
the FRDN process can be generated by a single qutrit---a quantum system with a three-dimensional Hilbert space~\cite{Fanizza24_Quantum}.
We find that neural networks trained on this classical stochastic process intrinsically learn the
finite-dimensional quantum generative mechanism, and represent Bayesian updates over the quantum state of this post-classical generator as the neural network observes more tokens during inference.
The observable alphabet only has two tokens $\mathcal{X} = \{ a, b \}$,
so the linear map from the latent space to the next-token distribution is non-invertible.



In Ref.~\cite{Riechers25_Identifiability}, we show that every 
stochastic process with a finite $d$-dimensional quantum representation has a GHMM representation with $d^2$ dimensions.
The minimal GHMM representation 
can be 
smaller, and can be obtained from any GHMM representation via the algorithm presented in Ref.~\cite{Uppe97a}.
In this case, 
as pointed out in Ref.~\cite{Fanizza24_Quantum},
the FRDN has a 4-state GHMM representation.  
In terms of parameters $\alpha \in \mathbb{R}$ and 
$0 < \lambda \leq 1/2$,
we can write the
matrices $T^{(a)}$ and $T^{(b)}$ 
as 
\begin{align}
T^{(a)} &= \dket{\omega} \! \dbra{\pi_0}
\end{align}
and 
\begin{align}
T^{(b)} &=
\lambda
\begin{bmatrix}
0&0&0&0 \\
0&1&0&0 \\
0&0&\cos \alpha & -\sin \alpha \\
0&0& \sin \alpha & \cos \alpha 
\end{bmatrix}
~,
\end{align}
where $\dket{\omega}^\top = \bigl[ 
1, \, 
1-\lambda, \, 1 + \lambda (\sin \alpha - \cos \alpha) , \, 
1 - \lambda (\sin \alpha + \cos \alpha)
\bigr]$
and $\dbra{\pi_0} = 
\bigl[ 1-\tfrac{1}{2(1-\lambda)} + \tfrac{1}{4}(c_+ + c_-), \, \tfrac{1}{2(1-\lambda)}, \, -c_+/4, \, -c_-/4 \bigr]
$
with $c_\pm \coloneqq \frac{1 - \lambda \cos \alpha \pm \lambda \sin \alpha}{(1 - \lambda \cos \alpha )^2 + \lambda^2 \sin^2 \alpha}$.\footnote{Note that there is a sine sign error in Eq.~(27) of Ref.~\cite{Fanizza24_Quantum} that we fix here.}

When $\pi/\alpha$ is irrational, there is no finite-dimensional HMM realization of the process.
When $\pi/\alpha$ is rational, there can still be a significant dimensional advantage. For the experiments shown in the appendix using the FRDN process, we used $\alpha=2000, \, \lambda=0.49$.

\subsection{Post-quantum: Moon process}

A minimal example of a post-quantum process---a classical stochastic process with a finite-dimensional GHMM generator, yet no finite HMM and no finite-dimensional quantum generator---has a simple three-dimensional representation, and an observable alphabet of three symbols $\mathcal{X} = \{ a, b, c \}$.
Following Ref.~\cite{Fanizza24_Quantum},
the linear maps generating the process are
\begin{align}
T^{(a)} = \nu \dket{m_0} \! \dbra{\mu_0} ~, \quad 
T^{(b)} = 
\nu
\begin{bmatrix}
\alpha &0&0 \\
0&1&0 \\
0& \ln \alpha & 1 
\end{bmatrix}
~, \; \text{and } \;
T^{(c)} = 
\nu
\begin{bmatrix}
\beta &0&0 \\
0 &1&0 \\
0& \ln \beta & 1 
\end{bmatrix} ~,
\end{align}
with $\alpha, \beta, \nu \in \mathbb{R}$,
such that $\alpha > 1 > \beta > 0$,
$\alpha+\beta \neq 2$,
$\ln(\alpha) / \ln(\beta) \in \mathbb{R} \setminus \mathbb{Q}$.
The scalar value
$\nu$ is chosen to make $T$ have a maximal eigenvalue of 1.

Following
Fanizza et al.~\cite{Fanizza24_Quantum}, in our experiments we use a parametrization of the post-quantum process with
$\alpha=e$, 
$\beta=1/2$,
$\dket{m_0} = \begin{bmatrix} 1 & 1 & 0\end{bmatrix}^\top$,
and 
$\dbra{\mu_0} = \begin{bmatrix} 1 & -1 & -1 \end{bmatrix}$.

\section{Experimental Methods for Training Networks}
\label{app:training-methods}

We performed standard self-supervised pretraining on next-token-prediction cross-entropy loss. Given the parameter vector of
weights and biases $\params$, and a given context $x_{1:\ell-1}$, a neural-network sequence model produces logits for the next token
$-\log \Pr_{\params}(X_\ell | X_{1:\ell-1} = x_{1:\ell-1})$.
Given an input sequence $x_{1:L}$, the pretraining loss is
$-\sum_{\ell=1}^{L-1} \log \Pr_{\params}(x_{\ell+1} |  x_{1:\ell})$.

\subsection{Experimental Design}

We conduct a comprehensive evaluation of four neural network architectures on four distinct stochastic processes, resulting in 16 experimental configurations. The architectures include transformers, LSTMs, GRUs, and vanilla RNNs, while the processes consist of Mess3 (classical), FRDN (quantum), Bloch Walk (quantum), and the Moon Process (post-quantum)---each representing different computational complexity classes as described in Appendix~\ref{app:ExampleProcesses}.

\subsection{Model Architectures}

For the Transformer architecture, we employ a 4-layer model implemented using the TransformerLens framework~\citep{nanda2022transformerlens}. The model uses multi-head attention with 4 heads (dimension 16 per head), 64-dimensional embeddings, and a 256-dimensional feed-forward network with ReLU activation. Layer normalization is applied before each sub-layer, and the model processes fixed sequences of 8 tokens with learned positional embeddings.

We compare three RNN variants, each configured with identical hyperparameters to ensure fair comparison: 4 recurrent layers with 64 hidden units per layer, unidirectional processing, one-hot input encoding, and a linear output projection. The variants differ only in their gating mechanisms: LSTM uses forget, input, and output gates; GRU employs reset and update gates; and the vanilla RNN uses simple tanh activation without gating.

\subsection{Training Methodology}

Training data is generated from each stochastic process with the following parameters (see Appendix~\ref{app:ExampleProcesses} for process definitions):
\begin{itemize}
    \item \textbf{Mess3}: $a=0.85$, $x=0.05$
    \item \textbf{Bloch Walk}: $\alpha=1$, $\beta=\sqrt{51}$
    \item \textbf{FRDN}: $\alpha=2000$, $\lambda=0.49$  
    \item \textbf{Moon Process}: $\alpha=e$, $\beta=1/2$
\end{itemize}

Each training example consists of an 8-token input sequence with corresponding next-token targets for each position. All experiments use consistent random seeding (seed=42) for reproducibility. We train using standard cross-entropy loss. Validation is performed every epoch on the full dataset, with model checkpoints saved every 100 epochs (201 total) and comprehensive metric logging via Weights \& Biases.

\subsubsection{Training Hyperparameters}

Table~\ref{tab:hyperparams} shows the core hyperparameters used across all experiments. All models are trained for 20,000 epochs using the Adam optimizer with learning rate $1 \times 10^{-4}$.

\begin{table}[h]
\centering
\caption{Core hyperparameters used across all experiments}
\label{tab:hyperparams}
\begin{tabular}{ll}
\toprule
\textbf{Hyperparameter} & \textbf{Value} \\
\midrule
Optimizer & Adam ($\beta_1=0.9$, $\beta_2=0.999$, $\epsilon=10^{-8}$) \\
Learning rate & $1 \times 10^{-4}$ \\
Weight decay & None \\
Gradient clipping & None \\
Epochs & 20,000 \\
Validation frequency & Every epoch \\
Checkpoint frequency & Every 100 epochs \\
Random seed & 42 \\
\bottomrule
\end{tabular}
\end{table}

The majority of experiments use the following configuration:

\begin{table}[H]
\centering
\caption{Standard configuration used by most experiments}
\label{tab:standard_config}
\begin{tabular}{ll}
\toprule
\textbf{Configuration} & \textbf{Standard Setting} \\
\midrule
Batch size & 128 \\
Batches per epoch & 200 \\
LR scheduler & ReduceLROnPlateau$^*$ \\
Total checkpoints & 201 \\
\bottomrule
\end{tabular}
\end{table}

\noindent $^*$ReduceLROnPlateau parameters: factor=0.5, patience=1000, cooldown=200, threshold=$10^{-6}$

While all RNN variants (LSTM, GRU, RNN) use the standard configuration above for all processes, we found that certain transformer experiments trained better with modified settings:

\begin{table}[h]
\centering
\caption{Experiment-specific variations from standard configuration}
\label{tab:variations}
\begin{tabular}{lccc}
\toprule
\textbf{Experiment} & \textbf{Batch Size} & \textbf{Batches/Epoch} & \textbf{LR Scheduler} \\
\midrule
Transformer-FRDN & 16 & 20 & None \\
Transformer-Moon & 16 & 20 & ReduceLROnPlateau \\
\bottomrule
\end{tabular}
\end{table}

\subsection{Implementation Details}

All experiments are implemented in PyTorch 2.0 with CUDA acceleration, using FP32 precision throughout. Training is distributed across multiple GPUs, with specific GPU assignments managed through a parallel execution framework. To ensure reproducibility, we use fixed random seeds.

 We maintain 4 layers across all architectures to ensure fair comparison of inductive biases rather than capacity differences. The high checkpoint frequency (every 100 epochs) enables detailed analysis of learning dynamics and convergence behavior. Finally, we deliberately avoid dropout, weight decay, or other regularization techniques to study the pure learning dynamics of each architecture on these processes. All code for training, analysis, and figure creation is publicly available, as well as saved model checkpoints and analysis results files, see Appendix~\ref{app:code} below.

\section{Analysis methods to probe for belief geometry in activations}
\label{app:methods-probe-beliefs}

\subsection{General Approach}
\label{app:LinRegression}

Our main analysis quantifies whether neural network activations encode belief states through an affine transformation of their internal activations. Given a neural activation vector $\act[w] \in \mathbb{R}^d$ (with e.g., activations from 
a particular position and
a single layer $d = d_\text{model}$, or 
activations from 
a particular position across all layers
$d = N d_\text{model}$)
induced by a token sequence $w \in \mathcal{X}^*$ and a proposed belief state $\boldsymbol{\eta}^{(w)} \in \mathbb{R}^{d_\text{g}}$, we seek an affine transformation:
\begin{align}
\boldsymbol{\eta}^{(w)} \approx \act[w] L + \boldsymbol{b} = \hat{\boldsymbol{\eta}}^{(w)}
\end{align}
where $L \in \mathbb{R}^{d \times d_\text{g}}$ is a linear map and $\boldsymbol{b} \in \mathbb{R}^{d_\text{g}}$ is a bias vector. We express this compactly using augmented notation:
\begin{align}
\boldsymbol{\eta}^{(w)} \approx \begin{bmatrix} 1 & \act[w] \end{bmatrix} \mathcal{L} = \hat{\boldsymbol{\eta}}^{(w)}
\end{align}
where $\mathcal{L} = \begin{bmatrix} \boldsymbol{b} \\ L \end{bmatrix} \in \mathbb{R}^{(1+d) \times d_\text{g}}$.

We assemble a dataset from anchor sequences $\mathcal{A} = \{w_n\}_{n=1}^{|\mathcal{A}|} \subset \mathcal{X}^*$. In our analysis, we take the set of anchor points to be all sequences generated by the process up to the context window length of the network. Let $A \in \mathbb{R}^{|\mathcal{A}| \times (1+d)}$ be the matrix of augmented activation vectors with $n$-th row $[1, \act[w_n]]$, and $\Gamma \in \mathbb{R}^{|\mathcal{A}| \times d_\text{g}}$ be the matrix with corresponding belief vector $\boldsymbol{\eta}^{(w_n)}$ in row $n$.

To reflect the process's true statistics, we weight each sequence by its probability 
$p_n = \frac{\dbra{\eta^{(\varnothing)}} T^{(w_n)} \dket{1}}{\sum_{w \in \mathcal{A}} \dbra{\eta^{(\varnothing)}} T^{(w)} \dket{1} }$
derived from the ground-truth generative model. 
In those cases where the anchor sequences consist of all allowable words up to some context length, i.e.,  $\mathcal{A} = \bigcup_{\ell=1}^{\ell_\text{context}} \{ w \in \Abet^\ell : \init T^{(w)} \dket{1} > 0\}$,
we will have 
$p_n = \dbra{\eta^{(\varnothing)}} T^{(w_n)} \dket{1} / \ell_\text{context}$.
The optimal transformation minimizes:
\begin{align}
\mathcal{L}^* = \argmin_{\mathcal{L}} \Bigl\langle \bigl\| \begin{bmatrix} 1 & \act[w_n] \end{bmatrix} \mathcal{L} - \boldsymbol{\eta}^{(w_n)} \bigr\|_2^2 \Bigr\rangle_{n}
\end{align}

Following the standard approach for weighted linear regression, we note that the solution via weighted least squares is:
\begin{align}
\mathcal{L}^* = (P^{1/2}A)^+ P^{1/2} \Gamma
\end{align}
where $P$ is a diagonal matrix with  $P_{nn} = p_n$, 
and $M^+$ denotes the regularized Moore--Penrose pseudoinverse of $M$.

\subsection{Implementation details}

\subsubsection{Deduplication}
Before regression, we identify and aggregate duplicate token prefixes. For each unique prefix, we retain the activation vector from its first occurrence, and then we sum the probabilities across all occurrences of the same prefix. This deduplication reduces computational cost while preserving the correct probability weighting.

\subsubsection{Computing the pseudoinverse}
For computational efficiency when evaluating multiple regularization parameters $r$, we compute the SVD once:
\begin{align}
P^{1/2}A = U \Sigma V^T ~,
\end{align}
where $\Sigma$ is a diagonal matrix whose diagonal elements are the singular values in descending order
$\Sigma_{n+1,n+1} \leq \Sigma_{n,n}$ for $n \geq 0$.
The regularized pseudoinverse for any $r>0$ is then
\begin{align}
(P^{1/2}A)^+_r = V\Sigma^+_r U^T ~,
\end{align}
where $\Sigma^+_r$ 
is diagonal with entries
\begin{align}
(\Sigma^+_r)_{n,n} = \begin{cases}
\frac{1}{\Sigma_{n,n}} & \text{if } \Sigma_{n,n} > r  \Sigma_{1,1} \\
0 & \text{otherwise} ~.
\end{cases} 
\end{align}

\subsubsection{Cross-Validation and Model Selection}

We use 10-fold cross-validation to select the optimal regularization parameter $r$ from a set combining $\{10^{-15}, 10^{-10}, 10^{-5}\}$ with 50 logarithmically-spaced values between $10^{-8}$ and $10^{-3}$.

For each fold and each candidate $r$:
\begin{enumerate}
\item Partition data into training (90\%) and validation (10\%) sets
\item Fit the weighted regression on the training set
\item Evaluate weighted error on the validation set: $\sum_i p_i \|\boldsymbol{\eta}^{(w_i)} - \hat{\boldsymbol{\eta}}^{(w_i)}\|_2$
\end{enumerate}

The $r$ minimizing average validation error across folds is selected for the final model trained on all data.

\subsubsection{Evaluation Metrics}

To quantify how well the belief states were represented in network activations through an affine map, we compute the root mean squared error, RMSE, of the fit, by first computing the mean squared error, MSE, according to:
\begin{align}
\text{MSE} &= \sum_i p_i \bigl\| \boldsymbol{\eta}^{(w_i)} - \hat{\boldsymbol{\eta}}^{(w_i)} \bigr\|_2^2 \\
\text{RMSE} &= \sqrt{\text{MSE}} ~.
\end{align}


\subsubsection{Cosine Similarity Analysis}

To assess whether the learned representations preserve the geometric relationships between belief states, we analyze pairwise cosine similarities. For a set of belief states $\{\boldsymbol{\eta}^{(w_i)}\}_{i=1}^{|\mathcal{A}|}$, we compute the cosine similarity matrix:
\begin{align}
S_{ij} = \frac{\boldsymbol{\eta}^{(w_i)} \cdot \boldsymbol{\eta}^{(w_j)}}{\|\boldsymbol{\eta}^{(w_i)}\| \|\boldsymbol{\eta}^{(w_j)}\|} ~.
\end{align}

We extract the upper triangle of this matrix (excluding the diagonal) to obtain a distribution of pairwise similarities. This analysis is performed for:
\begin{enumerate}
\item Ground truth belief states $\boldsymbol{\eta}^{(w_i)}$
\item Predicted belief states $\hat{\boldsymbol{\eta}}^{(w_i)}$ from the regression
\end{enumerate}

By comparing these geometric relationships, we evaluate whether the linear probe preserves the relative orientations between belief vectors. This provides a complementary view to the distance-based metrics, focusing on angular rather than Euclidean-distance relationships. The analysis is conducted separately for both Markov-order-3 approximations of the processes and full generator belief geometries.

\subsection{Control Experiments}

To verify that the learned representations are not artifacts of the architecture alone, we compare against networks with randomly initialized weights (i.e., before any training (backpropagation) has happened). This control undergoes the same regression analysis, allowing us to quantify how much structure arises from training versus architecture.

We also apply the same regression pipeline to classical Markov models of order 3, mapping their belief states to neural activations. We chose the Markov order to be 3 since in our main experimental condition for a post-classical process (the Bloch Walk Process), the Markov-order-3 approximation of the process has 64 states, and thus the corresponding 64-dimensional belief states should exactly fit into the hidden-state dimensionality of our neural networks. This provides a baseline for assessing whether neural networks learn representations aligned with classical approximations of the post-classical processes, or if they truly represent beliefs over the post-classical processes.

\section{Experimental Evidence for the universality of belief geometry in networks across architectures and types of stochastic processes}
\label{app:extended-figs}

Our central claim of universality—that the discovery of minimal belief geometry is independent of neural network architecture—is supported by a comprehensive set of experiments. While the main text highlights results for Transformers and LSTMs (Fig.~\ref{fig:main_results}), we performed identical analyses on vanilla Recurrent Neural Networks (RNNs) and Gated Recurrent Units (GRUs). The results, presented in Figs.~\ref{fig:transformer-extended},~\ref{fig:LSTM-extended},~\ref{fig:RNN-extended}, and \ref{fig:GRU-extended} below, show that all tested architectures successfully learn to linearly represent the correct classical, quantum, and post-quantum belief geometries of their respective training data. The affine-mapped activation geometries and their corresponding RMSE plots consistently demonstrate a strong preference for the minimal generator over classical approximations and random baselines, reinforcing that this phenomenon is a fundamental outcome of the training process rather than an artifact of a specific architecture.

\begin{figure}
    \centering
    \includegraphics[width=.85\linewidth]{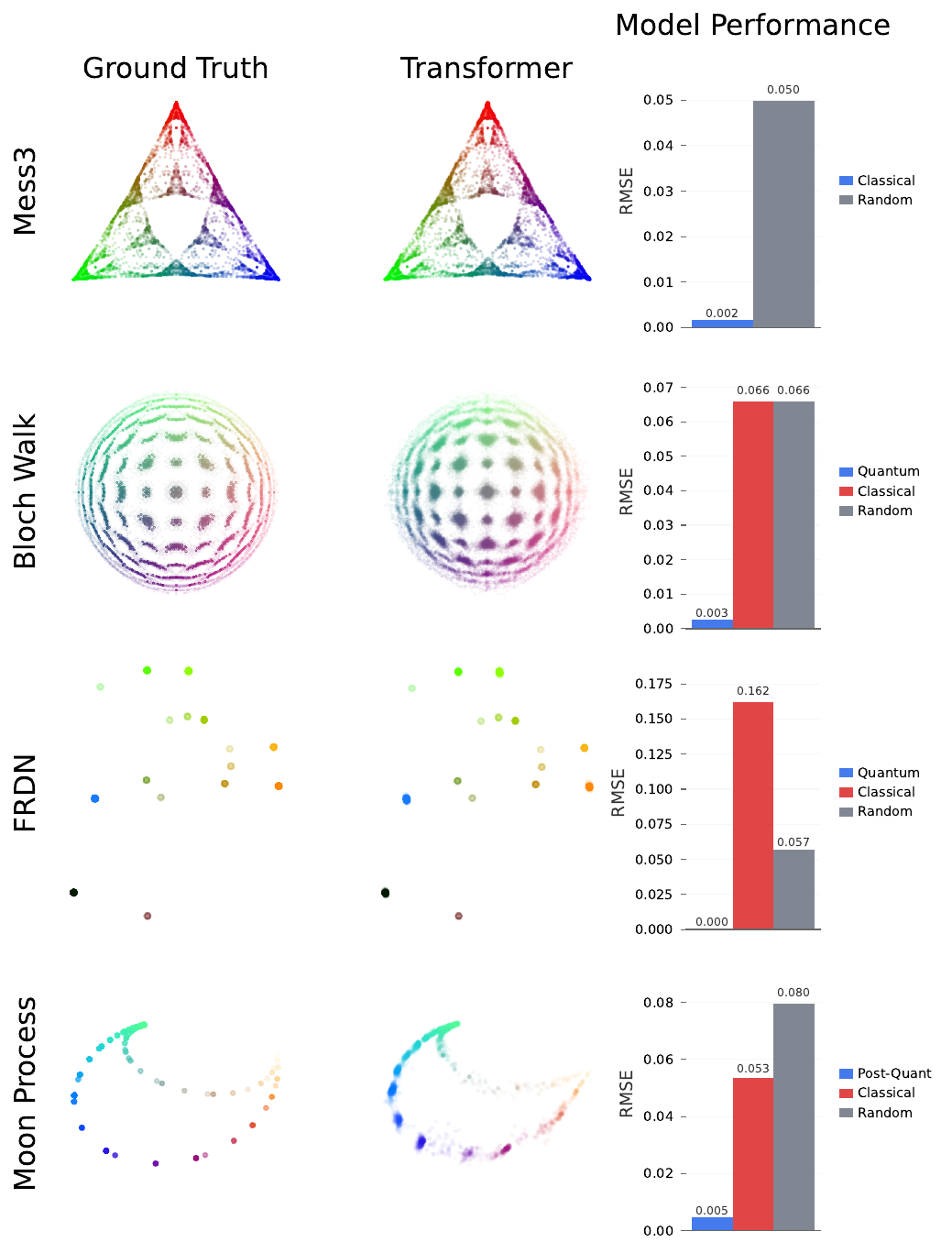}
    \caption{\textbf{Transformers learn the minimal belief geometry across all process classes.}
        Comparison of ground truth belief geometries (left column) with the affine-mapped activation geometries of a trained 4-layer Transformer (center column). The four rows correspond to the Mess3 (Classical), Bloch Walk (Quantum), FRDN (Quantum), and Moon (Post-Quantum) processes. Points are colored by their position in the ground truth space to visualize the preservation of geometric relationships. Bar plots (right column) show the root mean square error (RMSE) for fits to the true minimal generator (blue/purple), a classical Markov-order-3 approximation (red), and a randomly initialized network (gray). The Transformer consistently achieves low RMSE for the minimal generator, demonstrating its ability to learn the most compact representation, whether classical, quantum, or post-quantum.
    }
    \label{fig:transformer-extended}
\end{figure}

\begin{figure}
    \centering
    \includegraphics[width=.85\linewidth]{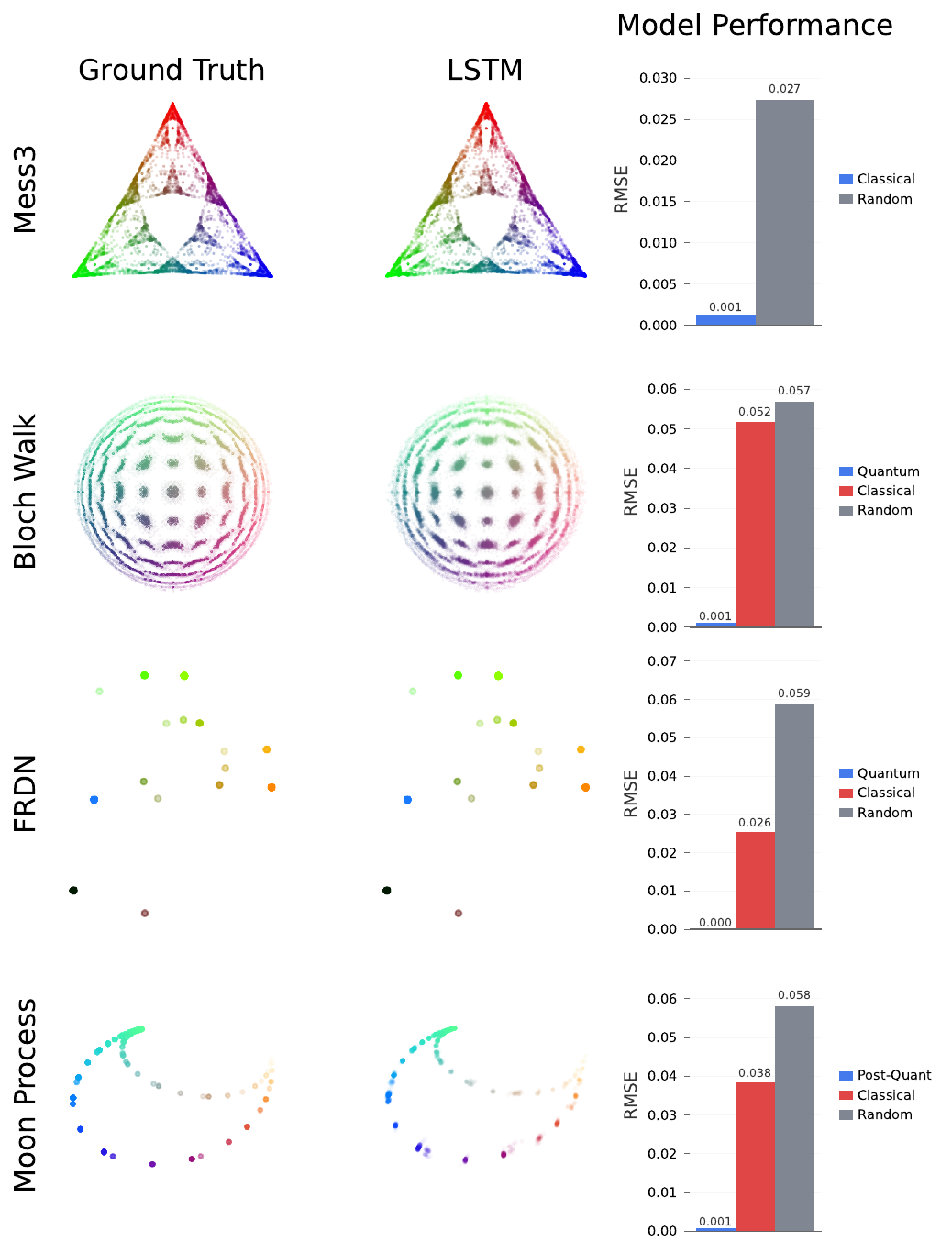}
     \caption{
        \textbf{LSTM architecture successfully discovers minimal belief geometries.}
        Comparison of ground truth belief geometries (left column) with the affine-mapped activation geometries from a trained 4-layer LSTM (center column). Each row presents a different stochastic process: Mess3 (Classical), Bloch Walk (Quantum), FRDN (Quantum), and Moon (Post-Quantum). The color correspondence illustrates that the LSTM preserves the relative structure of the belief space. The bar plots (right column) quantify the fit's accuracy, comparing the RMSE for the minimal generator (blue/purple) against a classical approximation (red) and a random network baseline (gray). The results demonstrate that, like Transformers, LSTMs effectively learn the correct classical, quantum, and post-quantum representations.
    }
    \label{fig:LSTM-extended}
\end{figure}

\begin{figure}
    \centering
    \includegraphics[width=.85\linewidth]{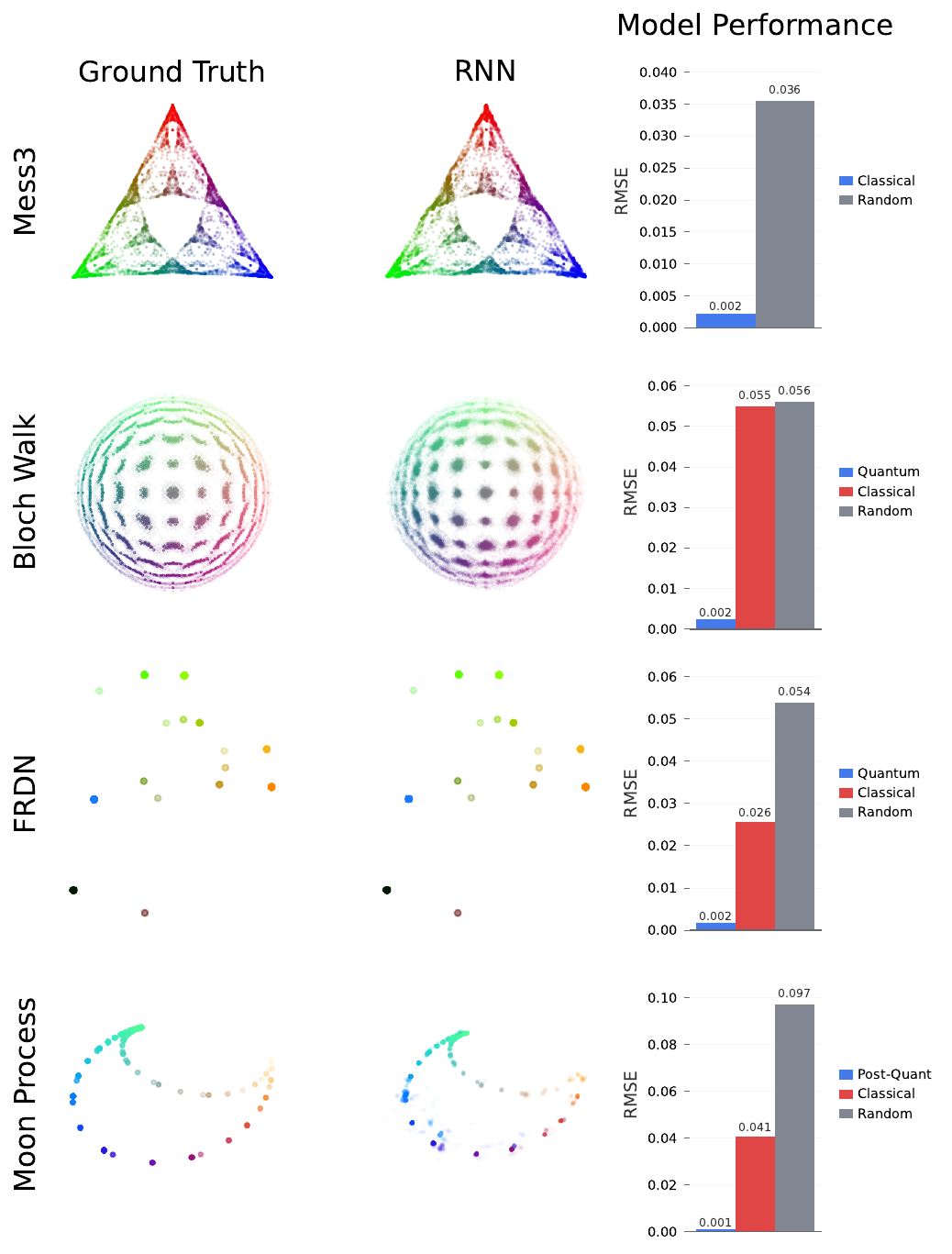}
    \caption{
        \textbf{Vanilla RNNs captures the correct minimal belief geometry.}
        Comparison of ground truth belief geometries (left column) with those learned by a trained 4-layer vanilla Recurrent Neural Network (RNN) (center column), for the Mess3, Bloch Walk, FRDN, and Moon processes. Despite its simpler architecture lacking gating mechanisms, the vanilla RNN learns to structure its activations in a way that linearly maps to the correct minimal belief geometry for classical, quantum, and post-quantum processes. The RMSE plots (right column) confirm a significantly better fit to the true generator (blue/purple) than to classical approximations (red) or random baselines (gray). This reinforces the universality of this phenomenon.
    }
    \label{fig:RNN-extended}
\end{figure}

\begin{figure}
    \centering
    \includegraphics[width=.85\linewidth]{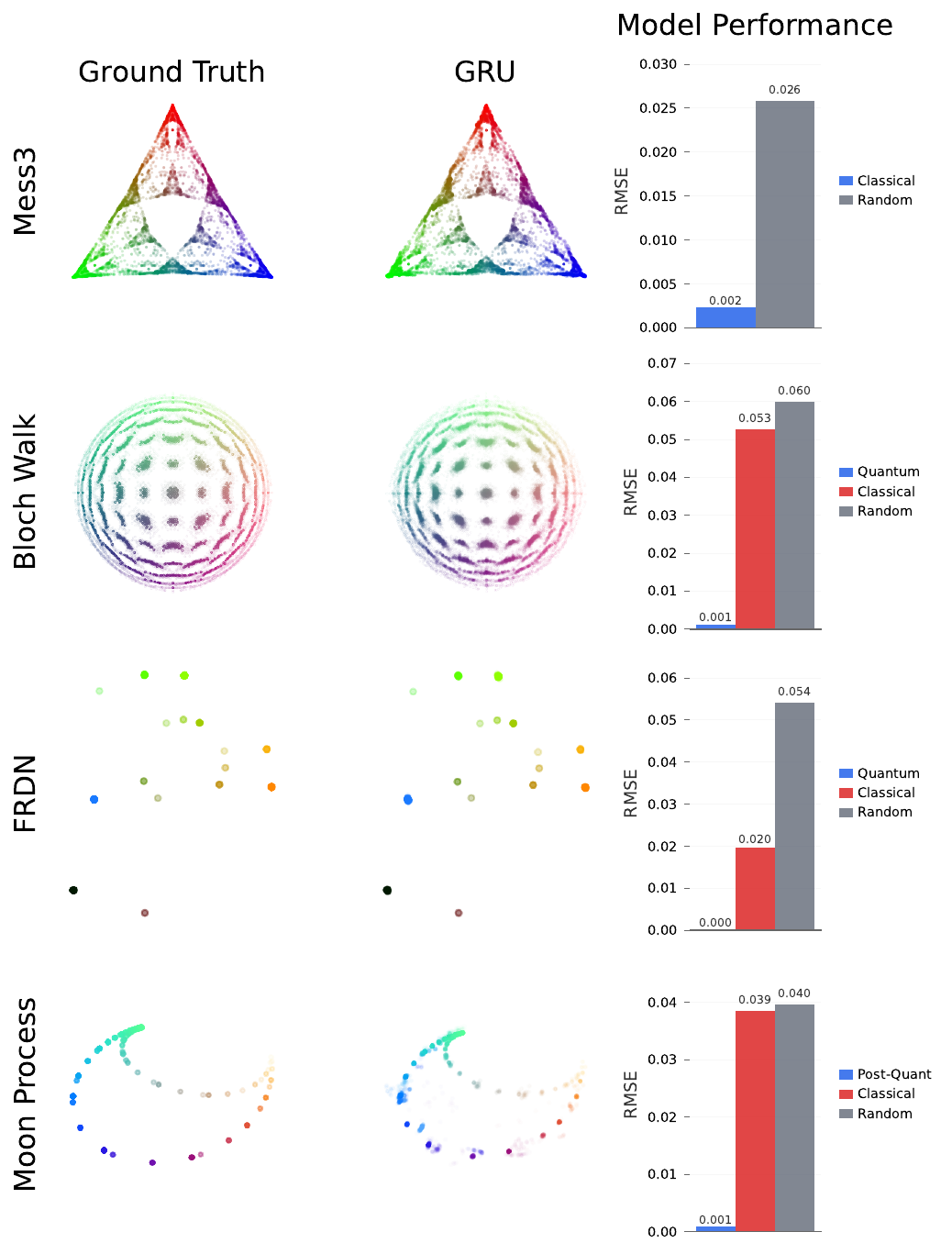}
    \caption{
        \textbf{GRU network performance mirrors other architectures in learning belief geometries.}
        Comparison of ground truth belief geometries (left column) with the affine-mapped activation geometries from a trained 4-layer Gated Recurrent Unit (GRU) network (center column) for all four process types. The visual and quantitative results are consistent with those from Transformer, LSTM, and vanilla RNN architectures. The GRU successfully identifies and represents the minimal belief geometry in its activations, as shown by the low RMSE values for the correct generator (blue/purple) compared to classical approximations (red) and random networks (gray). This further strengthens the claim that the discovery of minimal belief geometries is a universal feature of training recurrent-style networks on next-token prediction.
    }
    \label{fig:GRU-extended}
\end{figure}







\section{Code for Training and Analysis}
\label{app:code}

In the interest of reproducibility and completeness, we provide the codebase used to train networks, run analysis, and create the figures in this manuscript. That can be found \href{https://github.com/adamimos/epsilon-transformers/tree/quantum-public}{at this github link}.

We also include git commits containing the exact state of our code repository when we ran our experiments, links to configs, wandb logging during training, and a huggingface dataset that contains model checkpoints, configs, and regression analysis files. All available files are further explained in the following sections.

\subsection{Code Repository}

The code is publicly available at \href{https://github.com/adamimos/epsilon-transformers/tree/quantum-public}{https://github.com/adamimos/epsilon-transformers/tree/quantum-public}.  Please see the \href{https://github.com/adamimos/epsilon-transformers/blob/quantum-public/README.md}{README.md} for instructions on how to recreate all training and figure generation in this manuscript.

The repository contains:
\begin{itemize}
    \item Training scripts for all architectures (Transformer, LSTM, GRU, RNN)
    \item Activation analysis pipeline (\texttt{scripts/activation\_analysis/run\_regressions\_analysis.py})
    \item Figure generation scripts (\texttt{Fig2.py}, \texttt{Fig3.py}, \texttt{Fig4.py}, \texttt{FigAppendix.py})
    \item Process definitions and generators (\texttt{epsilon\_transformers/process/})
    \item Experiment configuration files (\texttt{scripts/experiment\_config\_*.yaml})
\end{itemize}

\subsection{Huggingface Dataset}

The complete dataset of trained model checkpoints and pre-computed analysis results is publicly available at \href{https://huggingface.co/datasets/SimplexAI/quantum-representations}{\texttt{SimplexAI/quantum-representations}}. The dataset contains:
\begin{itemize}
    \item \textbf{16 trained neural network models} (4 architectures $\times$ 4 processes)
    \item \textbf{Pre-computed belief state regression analysis results} for all models
    \item \textbf{Model checkpoints} at multiple training stages (201 checkpoints per model).
    \item \textbf{Training configurations and loss curves}
\end{itemize}

See the \href{https://huggingface.co/datasets/SimplexAI/quantum-representations/blob/main/README.md}{README.md} on Huggingface for more details.

\newcommand{\gitcommit}[1]{%
  \StrLeft{#1}{7}[\shortcommit]%
  \href{https://github.com/adamimos/epsilon-transformers/tree/#1}{\texttt{\shortcommit}}%
}

\subsection{Training Details and Data}

For completeness, we include links to the exact commits of the codebase used during training of each experiment in this manuscript, as well as links to Weights \& Biases for that training run, the training config, and the saved model checkpoints, in Table~\ref{tab:training-resources}. Note that losses in these files/links are reported normalized to the minimal possible loss given the process the network is trained on. Thus, minimum validation loss is 1.

\begin{table}[htbp]
\centering
\small
\caption{Training resources for all models. Each hyperlink provides access to the exact code version, training logs, configuration, and model checkpoints used in our experiments.}
\label{tab:training-resources}
\begin{tabular}{llccccccc}
\toprule
Process & Architecture & Sweep ID & Run ID & Code & W\&B & Config & Checkpoints \\
\midrule
\multirow{4}{*}{Mess3} 
    & Transformer & 20241205175736 & 23 & \gitcommit{93959eaed1c7d9316342703d816f40f390f1155c} & \href{https://wandb.ai/adamimos/quantum_transformer_20241205175736/runs/6c0crqcb}{[run]} & \href{https://huggingface.co/datasets/SimplexAI/quantum-representations/blob/main/models/20241205175736_23/run_config.yaml}{config} & \href{https://huggingface.co/datasets/SimplexAI/quantum-representations/tree/main/models/20241205175736_23}{HF} \\
    & LSTM & 20241121152808 & 55 & \gitcommit{434104d753915cea435eff07f03e376c29e6e20f} & \href{https://wandb.ai/adamimos/quantum_rnn_experiments_20241121152808/runs/8flc8p43?nw=nwuseradamimos}{[run]} & \href{https://huggingface.co/datasets/SimplexAI/quantum-representations/blob/main/models/20241121152808_55/run_config.yaml}{config} & \href{https://huggingface.co/datasets/SimplexAI/quantum-representations/tree/main/models/20241121152808_55}{HF} \\
    & GRU & 20241121152808 & 63 & \gitcommit{434104d753915cea435eff07f03e376c29e6e20f} & \href{https://wandb.ai/adamimos/quantum_rnn_experiments_20241121152808/runs/b87rrysv?nw=nwuseradamimos}{[run]} & \href{https://huggingface.co/datasets/SimplexAI/quantum-representations/blob/main/models/20241121152808_63/run_config.yaml}{config} & \href{https://huggingface.co/datasets/SimplexAI/quantum-representations/tree/main/models/20241121152808_63}{HF} \\
    & RNN & 20241121152808 & 71 & \gitcommit{434104d753915cea435eff07f03e376c29e6e20f} & \href{https://wandb.ai/adamimos/quantum_rnn_experiments_20241121152808/runs/z270i8fo}{[run]} & \href{https://huggingface.co/datasets/SimplexAI/quantum-representations/blob/main/models/20241121152808_71/run_config.yaml}{config} & \href{https://huggingface.co/datasets/SimplexAI/quantum-representations/tree/main/models/20241121152808_71}{HF} \\
\midrule
\multirow{4}{*}{\shortstack{Bloch\\Walk}} 
    & Transformer & 20241205175736 & 17 & \gitcommit{93959eaed1c7d9316342703d816f40f390f1155c} & \href{https://wandb.ai/adamimos/quantum_transformer_20241205175736/runs/4br1bez9}{[run]} & \href{https://huggingface.co/datasets/SimplexAI/quantum-representations/blob/main/models/20241205175736_17/run_config.yaml}{config} & \href{https://huggingface.co/datasets/SimplexAI/quantum-representations/tree/main/models/20241205175736_17}{HF} \\
    & LSTM & 20241121152808 & 49 & \gitcommit{434104d753915cea435eff07f03e376c29e6e20f} & \href{https://wandb.ai/adamimos/quantum_rnn_experiments_20241121152808/runs/jxl4ku8x?nw=nwuseradamimos}{[run]} & \href{https://huggingface.co/datasets/SimplexAI/quantum-representations/blob/main/models/20241121152808_49/run_config.yaml}{config} & \href{https://huggingface.co/datasets/SimplexAI/quantum-representations/tree/main/models/20241121152808_49}{HF} \\
    & GRU & 20241121152808 & 57 & \gitcommit{434104d753915cea435eff07f03e376c29e6e20f} & \href{https://wandb.ai/adamimos/quantum_rnn_experiments_20241121152808/runs/1s3qe4gv?nw=nwuseradamimos}{[run]} & \href{https://huggingface.co/datasets/SimplexAI/quantum-representations/blob/main/models/20241121152808_57/run_config.yaml}{config} & \href{https://huggingface.co/datasets/SimplexAI/quantum-representations/tree/main/models/20241121152808_57}{HF} \\
    & RNN & 20241121152808 & 65 & \gitcommit{434104d753915cea435eff07f03e376c29e6e20f} & \href{https://wandb.ai/adamimos/quantum_rnn_experiments_20241121152808/runs/x59y3sj2}{[run]} & \href{https://huggingface.co/datasets/SimplexAI/quantum-representations/blob/main/models/20241121152808_65/run_config.yaml}{config} & \href{https://huggingface.co/datasets/SimplexAI/quantum-representations/tree/main/models/20241121152808_65}{HF} \\
\midrule
\multirow{4}{*}{FRDN} 
    & Transformer & 20250422023003 & 1 & \gitcommit{a2fffc37cc80e652a3f407e539c8b3fa3a5215a0} & \href{https://wandb.ai/adamimos/quantum_transformer_20250422023003/runs/899g764w?nw=nwuseradamimos}{[run]} & \href{https://huggingface.co/datasets/SimplexAI/quantum-representations/blob/main/models/20250422023003_1/run_config.yaml}{config} & \href{https://huggingface.co/datasets/SimplexAI/quantum-representations/tree/main/models/20250422023003_1}{HF} \\
    & LSTM & 20241121152808 & 53 & \gitcommit{434104d753915cea435eff07f03e376c29e6e20f} & \href{https://wandb.ai/adamimos/quantum_rnn_experiments_20241121152808/runs/fl0fdqh2?nw=nwuseradamimos}{[run]} & \href{https://huggingface.co/datasets/SimplexAI/quantum-representations/blob/main/models/20241121152808_53/run_config.yaml}{config} & \href{https://huggingface.co/datasets/SimplexAI/quantum-representations/tree/main/models/20241121152808_53}{HF} \\
    & GRU & 20241121152808 & 61 & \gitcommit{434104d753915cea435eff07f03e376c29e6e20f} & \href{https://wandb.ai/adamimos/quantum_rnn_experiments_20241121152808/runs/joeft9du?nw=nwuseradamimos}{[run]} & \href{https://huggingface.co/datasets/SimplexAI/quantum-representations/blob/main/models/20241121152808_61/run_config.yaml}{config} & \href{https://huggingface.co/datasets/SimplexAI/quantum-representations/tree/main/models/20241121152808_61}{HF} \\
    & RNN & 20241121152808 & 69 & \gitcommit{434104d753915cea435eff07f03e376c29e6e20f} & \href{https://wandb.ai/adamimos/quantum_rnn_experiments_20241121152808/runs/zlgf8zsr}{[run]} & \href{https://huggingface.co/datasets/SimplexAI/quantum-representations/blob/main/models/20241121152808_69/run_config.yaml}{config} & \href{https://huggingface.co/datasets/SimplexAI/quantum-representations/tree/main/models/20241121152808_69}{HF} \\
\midrule
\multirow{4}{*}{\shortstack{Moon\\Process}} 
    & Transformer & 20250421221507 & 0 & \gitcommit{a2fffc37cc80e652a3f407e539c8b3fa3a5215a0} & \href{https://wandb.ai/adamimos/quantum_transformer_20250421221507/runs/pgfj928v?nw=nwuseradamimos}{[run]} & \href{https://huggingface.co/datasets/SimplexAI/quantum-representations/blob/main/models/20250421221507_0/run_config.yaml}{config} & \href{https://huggingface.co/datasets/SimplexAI/quantum-representations/tree/main/models/20250421221507_0}{HF} \\
    & LSTM & 20241121152808 & 48 & \gitcommit{434104d753915cea435eff07f03e376c29e6e20f} & \href{https://wandb.ai/adamimos/quantum_rnn_experiments_20241121152808/runs/zrbnc2hn?nw=nwuseradamimos}{[run]} & \href{https://huggingface.co/datasets/SimplexAI/quantum-representations/blob/main/models/20241121152808_48/run_config.yaml}{config} & \href{https://huggingface.co/datasets/SimplexAI/quantum-representations/tree/main/models/20241121152808_48}{HF} \\
    & GRU & 20241121152808 & 56 & \gitcommit{434104d753915cea435eff07f03e376c29e6e20f} & \href{https://wandb.ai/adamimos/quantum_rnn_experiments_20241121152808/runs/is6du73b?nw=nwuseradamimos}{[run]} & \href{https://huggingface.co/datasets/SimplexAI/quantum-representations/blob/main/models/20241121152808_56/run_config.yaml}{config} & \href{https://huggingface.co/datasets/SimplexAI/quantum-representations/tree/main/models/20241121152808_56}{HF} \\
    & RNN & 20241121152808 & 64 & \gitcommit{434104d753915cea435eff07f03e376c29e6e20f} & \href{https://wandb.ai/adamimos/quantum_rnn_experiments_20241121152808/runs/8yzz2ljf}{[run]} & \href{https://huggingface.co/datasets/SimplexAI/quantum-representations/blob/main/models/20241121152808_64/run_config.yaml}{config} & \href{https://huggingface.co/datasets/SimplexAI/quantum-representations/tree/main/models/20241121152808_64}{HF} \\
\bottomrule
\end{tabular}
\end{table}


\begin{thebibliography}{22}
	
\bibitem{Shai24_Transformers}
Adam~S. Shai, Sarah~E. Marzen, Lucas Teixeira, Alexander~Gietelink Oldenziel, and Paul~M. Riechers.
\newblock Transformers represent belief state geometry in their residual stream.
\newblock {\em NeurIPS, arXiv:2405.15943}, 2024.

\bibitem{Piotrowski25_Constrained}
Mateusz Piotrowski, Paul~M. Riechers, Daniel Filan, and Adam~S. Shai.
\newblock Constrained belief updates explain geometric structures in transformer representations.
\newblock {\em ICML}, 2025.

\bibitem{Riec19_Transforming}
P.~M. Riechers.
\newblock Transforming metastable memories: The nonequilibrium thermodynamics of computation.
\newblock In D.~Wolpert, C.~Kempes, P.~Stadler, and J.~Grochow, editors, {\em The Energetics of Computing in Life and Machines}, pages 353--380. SFI Press, 2019.

\bibitem{Nielsen10_Quantum}
Michael~A Nielsen and Isaac~L Chuang.
\newblock {\em Quantum computation and quantum information}.
\newblock Cambridge university press, 2010.

\bibitem{Plavala23_General}
Martin Pl{\'a}vala.
\newblock General probabilistic theories: {A}n introduction.
\newblock {\em Physics Reports}, 1033:1--64, 2023.

\bibitem{Fanizza24_Quantum}
Marco Fanizza, Josep Lumbreras, and Andreas Winter.
\newblock Quantum theory in finite dimension cannot explain every general process with finite memory.
\newblock {\em Communications in Mathematical Physics}, 405(2):50, 2024.

\bibitem{Monras16_Quantum}
Alex Monràs and Andreas Winter.
\newblock Quantum learning of classical stochastic processes: {T}he completely positive realization problem.
\newblock {\em Journal of Mathematical Physics}, 57(1):015219, 01 2016.

\bibitem{Uppe97a}
D.~R. Upper.
\newblock {\em Theory and Algorithms for Hidden {M}arkov Models and Generalized Hidden {M}arkov Models}.
\newblock PhD thesis, University of California, Berkeley, 1997.
\newblock {P}ublished by University Microfilms Intl, Ann Arbor, Michigan.

\bibitem{Balle15_Canonical}
Borja Balle, Prakash Panangaden, and Doina Precup.
\newblock A canonical form for weighted automata and applications to approximate minimization.
\newblock In {\em 2015 30th Annual ACM/IEEE Symposium on Logic in Computer Science}, pages 701--712. IEEE, 2015.

\bibitem{Riec18_SSAC1}
P.~M. Riechers and J.~P. Crutchfield.
\newblock Spectral simplicity of apparent complexity, {Part I}: {The} nondiagonalizable metadynamics of prediction.
\newblock {\em Chaos}, 28:033115, 2018.

\bibitem{Elliott22_Quantum}
Thomas~J Elliott, Mile Gu, Andrew~JP Garner, and Jayne Thompson.
\newblock Quantum adaptive agents with efficient long-term memories.
\newblock {\em Physical Review X}, 12(1):011007, 2022.

\bibitem{Zonnios25_Quantum}
Magdalini Zonnios, Alexander Boyd, and Felix Binder.
\newblock Quantum generation of stochastic processes: spectral invariants and memory bounds.
\newblock {\em New Journal of Physics}, 2025.

\bibitem{Gyamfi2020fundamentals}
Jerryman~A Gyamfi.
\newblock Fundamentals of quantum mechanics in {L}iouville space.
\newblock {\em European Journal of Physics}, 41(6):063002, 2020.

\bibitem{Riec24_Ideal}
Paul~M. Riechers, Chaitanya Gupta, Artemy Kolchinsky, and Mile Gu.
\newblock Thermodynamically ideal quantum state inputs to any device.
\newblock {\em PRX Quantum}, 5:030318, Jul 2024.

\bibitem{Riechers25_Identifiability}
Paul~M. Riechers and Thomas~J. Elliott.
\newblock Identifiability and minimality bounds of quantum and post-quantum models of classical stochastic processes.
\newblock {\em arXiv}, 2025.

\bibitem{Marz17a}
S.~E. Marzen and J.~P. Crutchfield.
\newblock Nearly maximally predictive features and their dimensions.
\newblock {\em Phys. Rev. E}, 95(5):051301(R), 2017.

\bibitem{Huh24_Position}
Minyoung Huh, Brian Cheung, Tongzhou Wang, and Phillip Isola.
\newblock Position: The platonic representation hypothesis.
\newblock In {\em Forty-first International Conference on Machine Learning}, 2024.

\bibitem{Crut10a}
J.~P. Crutchfield, C.~J. Ellison, J.~R. Mahoney, and R.~G. James.
\newblock Synchronization and control in intrinsic and designed computation: {An} information-theoretic analysis of competing models of stochastic computation.
\newblock {\em CHAOS}, 20(3):037105, 2010.
\newblock Santa Fe Institute Working Paper 10-08-015; arxiv.org:1007.5354 [cond-mat.stat-mech].

\bibitem{Pepper24_RNNs}
Keenan Pepper.
\newblock {RNNs} represent belief state geometry in their hidden states.
\newblock https://apartresearch.com/project/rnns-represent-belief-state-geometry-in-hidden-state, June 2024.
\newblock Research submission to the {C}omputational {M}echanics {H}ackathon research sprint co-hosted by Apart, PIBBSS, and Simplex.

\bibitem{Riechers25_Next}
Paul~M. Riechers, Henry~R. Bigelow, Eric~A. Alt, and Adam~S. Shai.
\newblock Next-token pretraining implies in-context learning.
\newblock {\em arXiv:2505.18373}, 2025.

\bibitem{Jako01}
L.~Jak{\'o}bczyk and M.~Siennicki.
\newblock Geometry of {B}loch vectors in two-qubit system.
\newblock {\em Physics Letters A}, 286(6):383--390, 2001.

\bibitem{nanda2022transformerlens}
Neel Nanda and Joseph Bloom.
\newblock Transformerlens.
\newblock \url{https://github.com/TransformerLensOrg/TransformerLens}, 2022.
	
\end{thebibliography}




\end{document}